\documentclass[11pt]{article}

\usepackage[preprint]{acl}

\usepackage[utf8]{inputenc} % allow utf-8 input
\usepackage[T1]{fontenc}    % use 8-bit T1 fonts
\usepackage{hyperref}       % hyperlinks
\usepackage{url}            % simple URL typesetting
\usepackage{booktabs}       % professional-quality tables
\usepackage{amsfonts}       % blackboard math symbols
\usepackage{nicefrac}       % compact symbols for 1/2, etc.
\usepackage{microtype}      % microtypography
\usepackage{xcolor}         % colors
\usepackage{makecell}
\usepackage{amsmath,amsfonts,amssymb,bm}
\usepackage{amsthm}
\usepackage{mathtools}
\usepackage{multirow}
\usepackage{tabularx}
\usepackage{array}

\usepackage{amsmath,amsfonts,bm}

\def\eqref#1{(\ref{#1})}
\DeclareMathAlphabet{\mathsfit}{\encodingdefault}{\sfdefault}{m}{sl}
\SetMathAlphabet{\mathsfit}{bold}{\encodingdefault}{\sfdefault}{bx}{n}

\newcommand{\eqdef}{\triangleq}

\newlength{\Oldarrayrulewidth}

\newcommand{\vocab}{\mathcal{V}}
\newcommand{\yreason}{\bm{y}^{\text{reason}}}
\newcommand{\yans}{\bm{y}^{\text{ans}}}
\newcommand{\ystar}{\bm{y}^*}
\newcommand{\ptrue}{p(\mathrm{True})}

\usepackage[most]{tcolorbox}

\newtcolorbox{findingbox}{
  colback=yellow!8,
  colframe=olive!60,
  boxrule=0.6pt,
  arc=0pt,
  left=6pt,
  right=6pt,
  top=3pt,
  bottom=3pt
}
\newtcolorbox{takeawaybox}{
  enhanced,
  breakable,
  colback=gray!4,
  colframe=gray!60,
  boxrule=0pt,
  leftrule=1.2pt,
  arc=0pt,
  left=4pt,
  right=4pt,
  top=3pt,
  bottom=3pt,
  before skip=4pt,
  after skip=4pt,
  fontupper=\small
}
\usepackage[normalem]{ulem}
\usepackage{enumitem}

\usepackage{realboxes}
\definecolor{success}{RGB}{40, 167, 69}

\newtcolorbox{promptbox}[1][]{
    colback=gray!5,
    colframe=gray!50,
    fonttitle=\bfseries,
    coltitle=black,
    enhanced,
    attach boxed title to top left={yshift=-2mm, xshift=2mm},
    boxed title style={colback=gray!20, colframe=gray!50},
    title=#1,
    sharp corners,
    boxrule=0.5pt,
    fontupper=\small\ttfamily,
    breakable
}

\usepackage[most]{tcolorbox}
\tcbuselibrary{listings}
\usepackage{listings}

\lstdefinestyle{promptstyle}{
  basicstyle=\ttfamily\small,
  breaklines=true,
  breakatwhitespace=false,
  columns=fullflexible,
  keepspaces=true,
  showstringspaces=false,
  aboveskip=0pt,
  belowskip=0pt
}

\usepackage{placeins}

\usepackage{times}
\usepackage{latexsym}

\usepackage[T1]{fontenc}
\usepackage[utf8]{inputenc}

\usepackage{microtype}

\usepackage{inconsolata}

\usepackage{graphicx}

\title{From token probabilities to calibrated confidence:\\An empirical study of mathematical question answering}

\author{Avery Ma\hspace{1.3em}Lorne Schell\hspace{1.3em}Vin Bhaskara\hspace{1.3em}Leila Pishdad \\\\[-0.6em]RBC Borealis}
\begin{document}
\maketitle
\begin{abstract}
Confidence estimation for large language models (LLMs) aims to estimate the probability that a generated answer is correct, while calibration aligns these estimates with empirical accuracy. 
Prior work has shown that token probabilities are often overconfident, we investigate whether these readily available signals can nevertheless provide well-calibrated confidence estimation for mathematical question answering.
% 
% Although token probabilities are readily available during inference, their effectiveness across different confidence estimation strategies remains under-explored. 
% Prior work has shown that token probabilities are often overconfident, but it remains unclear whether and how they can be transformed into useful confidence estimates.
% 
% In this work, we systematically study token-probability-based confidence estimators and post-hoc calibration for LLMs on mathematical reasoning tasks. 
% In this work, we present a systematic empirical study of token-probability-based confidence estimation for mathematical question answering.
% In this work, we present a systematic empirical study of token-probability-based confidence estimation and post-hoc calibration for mathematical question answering.
% Prior work has shown that token probabilities are often overconfident. In this work, we systematically investigate whether aggregation and post-hoc calibration can recover useful confidence estimates for mathematical question answering.
% 
We compare single-pass estimators, which reuse token probabilities from the original generation, with multi-pass estimators, which obtain additional confidence signals through verification or stochastic forward passes. 
While individual token probabilities can be highly saturated,
% We find that aggregating token probabilities over the full reasoning sequence produces more informative confidence estimates.
we find that aggregating token probabilities over the full sequence captures small but consistent differences between correct and incorrect generations, yielding more informative confidence estimates.
% we find that calibrated confidence arises from aggregating small but consistent probability differences across the reasoning trajectory.
% Despite their poor calibration when used directly, token probabilities yield informative confidence estimates when aggregated over the full reasoning sequence.
% 
% more reliable confidence signals emerge from token-level probability patterns aggregated across the full reasoning trajectory. 
% Across models and datasets, we find that probabilities associated with the final answer are often saturated and poorly calibrated, whereas aggregating token probabilities over the full reasoning sequence produces more informative confidence estimates.
% 
% Multi-pass methods generally yields more calibrated estimates. We study self-verification through model re-prompting, including a lower-cost in-situ variant, as well as Monte Carlo Dropout, which uses variation across stochastic forward passes.
Multi-pass methods can yield calibrated confidence estimates. 
We study two such approaches: self-verification through re-prompting, including a lower-cost in-situ variant, and Monte Carlo Dropout, which derives confidence from variation across stochastic forward passes.
% and introduce an efficient in-situ variant. 
% 
% We further demonstrate that confidence estimates derived from Monte Carlo Dropout samples achieve better calibration by capturing distributional uncertainty. 
% 
% Finally, we show that post-hoc calibration can recover useful signals from poorly calibrated estimators, with calibration data efficiency depending on dataset difficulty.
We further evaluate two post-hoc calibration methods, Platt scaling and isotonic regression, both of which substantially reduce in-domain calibration error. However, their data efficiency varies with dataset difficulty, and the calibration mappings often transfer asymmetrically across datasets and models.
\end{abstract}

\section{Introduction}
Large language models (LLMs) have achieved remarkable success on a broad range of complex, multi-step tasks~\citep{achiam2023gpt, team2023gemini}. As they are increasingly deployed in high-stakes domains such as financial forecasting~\citep{wu2023bloomberggpt, yu2023harnessing}, legal services~\citep{cai2025unilaw, luo2025automating}, and medical decision support~\citep{ravenda2025llms, bartels2025can}, generating a correct response alone is no longer sufficient. In these settings, it is equally important to know when a model's output can be trusted, motivating the need for accurate, well-calibrated estimates of response correctness.

A natural basis for confidence estimation is the token probabilities produced by LLMs during inference. However, \textit{raw} token probabilities often provide unreliable and saturated confidence signals~\citep{mielke2022reducing, kadavath2022language, chhikara2025mind}, because LLMs are trained to optimize next-token prediction~\citep{ouyang2022training, stiennon2020learning} rather than to estimate whether a completed response is correct. This raises a fundamental question: 
% 
% how can confidence signals be extracted reliably from token probabilities, and how can these signals be calibrated to better reflect actual correctness? 
whether token probabilities contain useful signals for confidence estimation and how effectively these signals can be calibrated to reflect empirical correctness.
% Addressing this question requires understanding not only which confidence estimators work well, but also why they work and how their reliability can be improved through post-hoc calibration.

In this work, we conduct an empirical study of token-probability-based confidence estimation and post-hoc calibration for mathematical question answering~\citep{cobbe2021training, patel2021nlp, gao2022pal}. We organize confidence estimators by the number of inference passes they require. For single-pass estimators, we analyze how confidence depends on the subset of generated tokens considered and the method used to aggregate their probabilities. For multi-pass methods, we study self-verification~\citep{kadavath2022language},  evaluate a lower-cost in-situ variant that appends the verification query directly to the original generation, and Monte Carlo (MC) Dropout~\citep{gal2016dropout}. Finally, we evaluate isotonic regression~\citep{zadrozny2002transforming} and Platt scaling~\citep{platt1999probabilistic}, examining how post-hoc calibration aligns confidence estimates with empirical accuracy. Our main contributions are:
\begin{itemize}[itemsep=-0.5mm, topsep=0mm, leftmargin=5mm]
    % \item We systematically analyze single-pass token-probability-based confidence estimators and show that final answer-token probabilities are highly saturated and poorly calibrated, making them unreliable for distinguishing correct from incorrect responses. In contrast, more reliable confidence signals emerge from aggregating weak but consistent token-level probability patterns across the full reasoning trajectory.
    \item We systematically compare single-pass confidence estimators derived from different subsets and aggregations of token probabilities. We find that the overconfidence issue documented in prior work is particularly pronounced for answer-token estimators. By contrast, aggregating probabilities over the full reasoning trajectory yields discriminative confidence estimates.

    % \item For multi-pass estimation, we introduce an efficient \textit{in-situ} self-verification variant that applies the verification query directly to the original generation trajectory. It matches the calibration of standard model re-prompting while reducing token-processing overhead by 88\%, demonstrating that verification-based confidence does not require re-encoding the full question--response pair, making it more practical for long reasoning outputs. 
    \item For multi-pass estimation, we evaluate an efficient \textit{in-situ} self-verification variant that appends the verification query directly to the original response. It achieves calibration comparable to standard model re-prompting while reducing token-processing overhead by 88\%, indicating that verification-based confidence can be estimated without re-encoding the full question--response pair and may therefore be particularly efficient for long reasoning outputs.

    % \item We further analyze multi-pass confidence estimation using MC Dropout. We show that dropout-induced variation in answer-token predictive distributions provides reliable confidence signals even when the discrete decoded output remains unchanged.
    \item We further analyze MC Dropout for multi-pass confidence estimation. We find that variation in answer-token predictive distributions captures uncertainty not reflected in the decoded output, even when the generated answer remains unchanged across stochastic forward passes.

    % \item We show that post-hoc calibration can recover useful confidence signals from heavily miscalibrated estimators. In particular, isotonic regression substantially improves calibration, often with as few as 50 examples, and calibration efficiency depends strongly on dataset difficulty.
    \item We evaluate Platt scaling and isotonic regression and find that both generally reduce in-domain calibration error, with isotonic regression often effective using as few as 50 examples.  Together with our cross-domain analysis, these results show that calibration efficiency depends on dataset difficulty, while transfer depends on the specific source and target model--dataset pairs.
\end{itemize}
\section{Related Work}\label{sec:related_work}
\textbf{Confidence estimation methods for LLMs} are typically categorized as white-box or black-box~\citep{geng2024survey}. 
Black-box methods derive confidence from generated outputs, including verbalized confidence~\citep{xiong2023can} and consistency across sampled generations~\citep{lyu2025calibrating}. 
White-box methods instead leverage internal representations ~\citep{lin2024contextualized} or logit-based signals~\citep{huang2023look,vazhentsev2023efficient,kuhn2023semantic}. 
Token probabilities are particularly appealing because they are available during inference and lie in $[0,1]$, making them natural candidates for confidence scores. 
However, they are often overconfident, concentrated near one, and poorly calibrated for answer correctness~\citep{mielke2022reducing,chhikara2025mind}. Motivated by these limitations, we examine how token selection, aggregation, and post-hoc calibration affect their usefulness for confidence estimation.

Multi-pass methods obtain additional confidence signals through repeated forward passes. The $\ptrue$ approach re-prompts the model to judge its generated answer and computes confidence by normalizing the probabilities of \texttt{True} and \texttt{False}~\citep{kadavath2022language}. This requires processing the full question--answer context again.

MC Dropout estimates uncertainty from variation across stochastic forward passes~\citep{gal2016dropout} and has been studied in classification~\citep{ficsor2025sue}, summarization~\citep{zablotskaia2023uncertainty}, and generative QA~\citep{mora2024uncertainty}. More recently, \citet{zhang2026tokur} estimate token-level uncertainty conditioned on a
fixed reasoning response using weight perturbations and aggregate it over
the full sequence.
We instead apply standard MC Dropout to a fixed model-generated reasoning trajectory and measure variation in final-answer-token distributions, evaluating whether it yields calibrated estimates of answer correctness.

\textbf{Confidence calibration} aims to align confidence scores with empirical correctness. Recent work has explored increasingly complex calibration methods for LLMs, including auxiliary LLM-based calibrators~\citep{tan2026basecal, ulmer2024calibrating}, layer-specific calibration~\citep{joshi2025calibration, stolfo2024confidence}, and fine-tuning-based approaches~\cite{li2026conftuner, lin2022teaching}. 
In contrast, we revisit two classic post-hoc approaches, Platt scaling~\citep{platt1999probabilistic} and isotonic regression~\citep{zadrozny2002transforming}, and show that they can substantially improve token-probability-based confidence estimates. Their simplicity lets us study in-domain and cross-domain calibration, along with calibration-data efficiency. We describe calibration metrics in Section~\ref{sec:calibration}.

\section{Preliminaries}
In this section, we introduce the notation and problem setting for our study of confidence estimation and calibration in LLMs.

\subsection{LLM Inference}\label{sec:llm_inference}
LLMs are 
% transformer-based 
neural networks that model sequential data by predicting the next token in a sequence~\citep{vaswani2017attention}. 
% \vin{can we drop "are transformer-based neural networks that" to keep it architecture-agnostic?\fixed} 
In this work, we focus on LLMs trained for text generation~\citep{brown2020language, achiam2023gpt, touvron2023llama}.

Let $\vocab$ denote a finite vocabulary. Given an input sequence $\bm{x} = (x_1, \dotsc, x_m) \in \vocab^m$, an LLM defines a probability distribution $\mathbb{P}(\cdot \mid \bm{x}) \in \Delta(\vocab)$ over the next token, where $\Delta(\vocab)$ denotes the probability simplex over $\vocab$. For an output sequence $\bm{y} = (y_1,\dotsc,y_n) \in \vocab^{n}$, the autoregressive factorization is given by
$
\mathbb{P}(\bm{y} \mid \bm{x})
=
\prod_{i=1}^{n} \mathbb{P}(y_i \mid \bm{x}, \bm{y}_{<i}),
$
where $\bm{y}_{<i} = (y_1,\dotsc,y_{i-1})$.

% We model the autoregressive behavior of LLMs by defining a distribution over sequences $\bm{y} = (y_1,\dotsc,y_n) \in \vocab^{n}$, where $n \in \mathbb{N}_+$, and factorizing it via the chain rule: $p(y_1, \dotsc, y_n \mid \bm{x}) = \prod_{i=1}^{n} p(y_i \mid \bm{x}, y_{<i})$.

% \subsection{Structured Outputs in Reasoning Tasks}
\subsection{Mathematical Question Answering with Verifiable Final Answers}
% \subsection{Structured Outputs in Mathematical Question Answering}
Confidence estimation requires a notion of output correctness. For open-ended dialogue~\citep{li2017dailydialog}, correctness can be ambiguous and task-dependent, making confidence difficult to define and evaluate~\citep{liu2023g, mendoncca2024benchmarking}. We therefore focus on mathematical question answering~\citep{cobbe2021training, patel2021nlp, gao2022pal}, 
which provides a \emph{structured} and \emph{verifiable} setting: model outputs typically contain multi-step reasoning followed by a final answer~\citep{wei2022chain}, and correctness can be evaluated against a ground-truth answer.

We decompose each generated sequence as
$
\bm{y} = (\yreason, \yans),
$
where $\yreason$ denotes the intermediate reasoning tokens and $\yans$ denotes the final answer tokens. We refer to $\yans$ as the \textit{answer tokens}. Given a ground-truth answer $\ystar$, we define a correctness function
$
\mathbb{I}(\yans, \ystar) \in \{0, 1\}
$,
which evaluates correctness under any chosen metric (e.g., exact match, semantic equivalence~\citep{peinelt2020tbert}, or LLM-as-a-Judge~\citep{perez2022red}).

% Although we focus on mathematical reasoning datasets, the calibration challenges studied here are relevant to other tasks with structured and verifiable outputs. Token-level probabilities are available throughout both reasoning steps and final answers. While prior work has leveraged various aggregation strategies~\citep{lin2024contextualized, gupta2024language, huang2024calibrating, jiang2020can, huang2023look, orgad2024llms}, these choices are often treated as baseline comparisons, with evaluation focused on uncertainty-based error detection rather than calibrated confidence estimation. As a result, it remains unclear how token probabilities should be aggregated to produce confidence scores that reflect empirical correctness.
Because token probabilities are available for both the reasoning trajectory and the final answer during inference, confidence can be constructed from different token subsets and aggregation rules. Prior work has considered several such choices~\citep{lin2024contextualized,gupta2024language, huang2024calibrating,jiang2020can,huang2023look,orgad2024llms}, often as baselines for error detection. We instead examine how these choices affect both the discrimination and calibration of confidence estimates for mathematical question answering.
\section{Confidence Estimation}\label{sec:estimation}
Our goal is to systematically study confidence estimators derived from token probabilities and their calibration behavior. We distinguish confidence estimation from calibration: confidence estimation constructs a scalar score to reflect answer correctness, while calibration adjusts this score to better align with empirical accuracy. In this section, we focus on the estimation step.

\subsection{Problem Formulation}
Confidence is commonly interpreted as the probability that a model's output is correct~\citep{guo2017calibration}. Consider a generated response $\hat{\bm{y}}$. A confidence estimator $c(\bm{x}, \hat{\bm{y}})$ aims to approximate the probability that the generated answer is correct: 
$
c(\bm{x}, \hat{\bm{y}}) 
\eqdef
\mathbb{P}\big(\mathbb{I}(\bm{y}^{\text{ans}}, \bm{y}^*) = 1 \mid \bm{x}, \hat{\bm{y}}\big)
$. 
This confidence score $c$ is different from the sequence likelihood $\mathbb{P}(\bm{y} \mid \bm{x})$ defined in Section~\ref{sec:llm_inference}. 
These quantities are often not the same: answer tokens can be highly likely when conditioned on an incorrect reasoning trajectory, while low-probability tokens along the trajectory may signal potential errors, as we later show in Section~\ref{sec:exp}.
Thus, estimating confidence from token probabilities requires two choices: which generated tokens to consider and how to aggregate their probabilities.

\subsection{Single-pass Estimation}\label{sec:single-pass}
We group confidence estimators based on the number of forward passes required. Single-pass estimators reuse token probabilities from the original generation and therefore incur no additional forward passes.
% Motivated by the structured output format of mathematical question answering, we study two design choices: (i) which generated tokens are included and (ii) how their probabilities are aggregated. 
% 
% We instantiate the two choices above using either the full generated sequence 
% For the token subset, we consider either the full generated sequence 
% $(\yreason, \yans)$ or only the answer tokens $\yans$. For aggregation, 
% we consider length-normalized joint probability~\citep{wu2016google, malinin2020uncertainty} and average probability~\citep{huang2023look, jiang2020can, orgad2024llms}. 
% we consider the arithmetic mean of token probabilities~\citep{huang2023look,jiang2020can,orgad2024llms} and the length-normalized joint probability, equivalent to the geometric mean of token probabilities
% ~\citep{wu2016google,malinin2020uncertainty}.
We instantiate the two choices above using either the full generated sequence $(\yreason,\yans)$ or only the answer tokens $\yans$, and aggregating their probabilities using either the arithmetic mean~\citep{huang2023look,jiang2020can,orgad2024llms} or the length-normalized joint probability, equivalent to the geometric mean of token probabilities~\citep{wu2016google,malinin2020uncertainty}.
% ~\citep{wu2016google,malinin2020uncertainty}.
% This yields four estimators: sequence joint (Seq Joint), sequence average (Seq Avg), answer joint (Ans Joint), and answer average (Ans Avg). 
Combining these choices yields four estimators: sequence joint (\textsc{Seq Joint}), sequence average (\textsc{Seq Avg}), answer joint (\textsc{Ans Joint}), and answer average (\textsc{Ans Avg}).
% Following prior work~\citep{wu2016google, malinin2020uncertainty}, the joint variants use length normalization to mitigate bias toward shorter sequences. Estimation based on these methods is very efficient, as it introduces no additional cost beyond standard inference. 
Length normalization reduces the joint probability's inherent preference for shorter sequences. 

\begin{table}[t]
\centering
\scriptsize
\renewcommand{\arraystretch}{1.5}
\setlength{\tabcolsep}{12pt}
% \small
% \renewcommand{\arraystretch}{1.25}
% \setlength{\tabcolsep}{8pt}

\begin{tabular}{llp{6.5cm}}
\toprule
\textbf{Method} & \textbf{Definition} \\
\midrule

\textsc{Seq Joint}
& $\left(\prod_{i=1}^{|\hat{\bm{y}}|} \mathbb{P}(y_i \mid \bm{x}, \hat{\bm{y}}_{<i})\right)^{1/|\hat{\bm{y}}|}$ \\

\textsc{Seq Avg}
& $\frac{1}{|\hat{\bm{y}}|}\sum_{i=1}^{|\hat{\bm{y}}|}\mathbb{P}(y_i \mid \bm{x}, \hat{\bm{y}}_{<i})$ \\

\textsc{Ans Joint}
& $\left(\prod_{j=1}^{|\yans|}\mathbb{P}(y^{\mathrm{ans}}_j \mid \bm{x}, \yreason, \yans_{<j})\right)^{1/|\yans|}$ \\

\textsc{Ans Avg}
& $\frac{1}{|\yans|}\sum_{j=1}^{|\yans|}\mathbb{P}(y^{\mathrm{ans}}_j \mid \bm{x}, \yreason, \yans_{<j})$ \\
\midrule

$\ptrue$
& $\frac{\mathbb{P}(y_{\mathtt{True}}\mid \bm{x}^{\mathrm{ver}})}
{\mathbb{P}(y_{\mathtt{True}}\mid \bm{x}^{\mathrm{ver}})+\mathbb{P}(y_{\mathtt{False}}\mid \bm{x}^{\mathrm{ver}})}$ \\

$c_{\mathrm{BALD}}$
& $\frac{1}{|\yans|}\sum_{j=1}^{|\yans|}\phi(\mathrm{BALD}_j)$ \\

\bottomrule
\end{tabular}
\caption{\
\textbf{Confidence estimation using token probabilities.}
Single-pass estimators use token probabilities from the original generation: sequence-based estimators aggregate probabilities over the full response, whereas answer-based estimators use the final-answer tokens. Multi-pass estimators obtain additional confidence signals through $\ptrue$-based self-verification~\citep{kadavath2022language} or $c_{\mathrm{BALD}}$ computed from stochastic forward passes~\citep{houlsby2011bayesian}.
}
\label{tab:estimators}
\end{table}

\subsection{Multi-pass Estimation}\label{sec:multi-pass}
% The single-pass estimators above compute confidence using only token probabilities obtained during the original generation. We next consider two multi-pass approaches that use additional forward passes to obtain token probabilities beyond the original output.
The single-pass estimators above use only token probabilities produced during the original generation. We next consider two multi-pass approaches that obtain additional confidence signals through repeated forward passes.

\textbf{Self-verification via re-prompting:} 
% Prior work has shown that language models can estimate their own confidence through verification-style re-prompting~\citep{kadavath2022language}.
% After generating the original response, the model is queried again with a correctness-verification prompt $\bm{x}^{\mathrm{ver}} = g(\bm{x}, \hat{\bm{y}})$ constructed from the input and generated response.
% As shown in Table~\ref{tab:estimators}, $\ptrue$ computes confidence by normalizing the probabilities assigned to the \texttt{True} and \texttt{False} tokens.
Prior work has used verification-style prompting to elicit confidence from language models~\citep{kadavath2022language}. After generating a response, the model is queried with a verification prompt
$\bm{x}^{\mathrm{ver}} = g(\bm{x},\hat{\bm{y}})$ constructed from the original input and generated response. As shown in Table~\ref{tab:estimators}, $\ptrue$ computes confidence by normalizing the probabilities assigned to
the \texttt{True} and \texttt{False} tokens.
% While $\ptrue$ is an effective and widely used baseline~\citep{orgad2024llms, lin2024contextualized, mahaut2024factual, kuhn2023semantic, lyu2025calibrating}, it requires an additional forward pass over the re-prompted question--answer pair. As a result, its computational overhead grows with response length, which can be substantial for practical applications involving long reasoning trajectories.
% We later study why $\ptrue$ is better calibrated than answer-token estimators in Section~\ref{sec:exp}, and use this insight to design an in-situ variant that preserves the self-verification setting while avoiding a full re-prompt.
Although $\ptrue$ is a widely used baseline~\citep{orgad2024llms,lin2024contextualized,mahaut2024factual,lyu2025calibrating}, it requires processing the question--response context again. Its token-processing overhead therefore increases with response length and can become substantial for long reasoning trajectories. In Section~\ref{sec:exp}, we evaluate the calibration of $\ptrue$ and a simple \textit{in-situ} variant that appends the verification prompt to the original generation trajectory, avoiding re-encoding the full context.

\textbf{Confidence estimation via MC Dropout:} 
% Lastly, we consider a multi-sample estimator based on MC Dropout, a widely used approximate Bayesian method for estimating model uncertainty~\citep{gal2016dropout, gal2017uncertainty}.
% While MC Dropout can be incorporated into estimation in multiple ways~\citep{malinin2020uncertainty, gao2025flue}, 
% we propose to estimate distributional uncertainty at the answer-token positions along a fixed trajectory. Given the model's initial response, we evaluate whether the model remains confident in its answer across dropout-induced stochastic forward passes. If the model has internal support for its answer conditioned on its own reasoning trajectory~\citep{kadavath2022language, yin2023large, jiang2021can}, we should expect different stochastic passes to produce similar predictive distributions at the answer-token positions, whereas larger variation across these distributions indicates lower confidence.
We next evaluate MC Dropout, a widely used approximate Bayesian approach for estimating uncertainty through stochastic forward passes~\citep{gal2016dropout,gal2017uncertainty}. Prior work has applied MC dropout to uncertainty estimation in language models across several settings~\citep{gao2025flue,zhang2026tokur}. 
Our analysis measures variation in the predictive distributions of answer tokens conditioned on a fixed reasoning trajectory. This design avoids the cost of repeatedly generating complete alternative reasoning trajectories, while leveraging the structured outputs of mathematical question answering, where a reasoning trajectory is followed by a compact, verifiable final answer.

We first generate a response with dropout disabled, then fix the trajectory $(\yreason,\yans)$ and enable dropout during inference. Importantly, we do not regenerate the response under each dropout mask. Instead, for each stochastic forward pass $t=1,\dotsc,T$ and answer-token position $j$, we evaluate the next-token predictive distribution conditioned on the same fixed prefix:
% For each stochastic forward pass $t = 1,\dotsc,T$ and answer-token position $j$, we evaluate the next-token predictive distribution conditioned on the fixed prefix:
% $
% \mathbb{P}_t(\cdot \mid \bm{x}, \yreason, \bm{y}^{\mathrm{ans}}_{<j}).
% $
% where each pass corresponds to a different dropout mask.
% \vin{What about models that don't have dropout? Can we comment about it here that we inject dropout layers (where specifically? the MLP layers of the transformer or attention or etc?)?}
% \am{i talk about it in the exp implementation details} %ah perfect! thanks np
% \[
$
\mathbb{P}_{t,j}(\cdot)
\coloneqq
\mathbb{P}_{t}
\left(
\cdot
\mid
\bm{x},\yreason,\yans_{<j}
\right)$.
% \]
If the prediction at an answer position is stable under dropout, these distributions should be similar across stochastic
passes. Greater disagreement indicates higher uncertainty about that answer
token given the generated reasoning trajectory. Let
% \[
$
\bar{\mathbb{P}}_j(\cdot)
\coloneqq
\frac{1}{T}
\sum_{t=1}^{T}
\mathbb{P}_{t,j}(\cdot)
$
% \]
denote the predictive distribution averaged across dropout samples.
We quantify disagreement at answer-token position $j$ using Bayesian Active
Learning by Disagreement (BALD)~\citep{houlsby2011bayesian, atighehchian2019baal}:
% We quantify variation in the predictive distributions across the answer, we compute the Bayesian Active Learning by Disagreement (BALD) score. 
% For each $j = 1,\dotsc,|\yans|$, we define
% \begin{align}
% \mathrm{BALD}_j
% &=
% H\!\left(Y^{\mathrm{ans}}_j \mid \bm{x}, \yreason, \bm{y}^{\mathrm{ans}}_{<j}\right)
% -
% \mathbb{E}_{\theta}
% \left[
% H\!\left(Y^{\mathrm{ans}}_j \mid \bm{x}, \yreason, \bm{y}^{\mathrm{ans}}_{<j}, \theta\right)
% \right] \nonumber \\
% &=
% \underbrace{
% H\!\left(
% \frac{1}{T} \sum_{t=1}^T
% \mathbb{P}_t(\cdot \mid \bm{x}, \yreason, \bm{y}^{\mathrm{ans}}_{<j})
% \right)
% }_{\text{Predictive entropy}}
% -
% \underbrace{
% \frac{1}{T} \sum_{t=1}^T
% H\!\left(
% \mathbb{P}_t(\cdot \mid \bm{x}, \yreason, \bm{y}^{\mathrm{ans}}_{<j})
% \right)
% }_{\text{Expected conditional entropy}},
% \label{eq:bald}
% \end{align}

% Let 
% $
% \mathbb{P}_{t,j}(\cdot) \coloneqq 
% \mathbb{P}_t(\cdot \mid \bm{x}, \yreason, \yans_{<j})
% $
% denote the predictive distribution at answer-token position $j$ under dropout sample $t$, and let
% $
% \bar{\mathbb{P}}_j(\cdot) \coloneqq 
% \frac{1}{T}\sum_{t=1}^{T}\mathbb{P}_{t,j}(\cdot)
% $
% denote the averaged predictive distribution. For each $j = 1,\dotsc,|\yans|$, we define the BALD score as
\vspace{-5mm}
\begin{equation}
\mathrm{BALD}_j
=
\underbrace{
H\!\left(\bar{\mathbb{P}}_j\right)
}_{\text{predictive entropy}}
-
\underbrace{
\frac{1}{T}
\sum_{t=1}^{T}
H\!\left(\mathbb{P}_{t,j}\right)
}_{\substack{\text{expected conditional}\\\text{entropy}}}.
\label{eq:bald}
\end{equation}
\vspace{-5mm}

The first term measures the entropy of the averaged predictive distribution across dropout samples, while the second term measures the average entropy within individual samples. A higher BALD score indicates greater variation among predictive distributions at the answer-token position, suggesting lower confidence in the answer conditioned on the fixed generated trajectory.

Because BALD is an uncertainty score rather than a probability of
correctness, we convert each token-level score into a confidence score using
a fixed monotone decreasing transformation
$\phi:\mathbb{R}\rightarrow[0,1]$ and average across answer-token
positions:
$
c_{\mathrm{BALD}}
=
\frac{1}{|\yans|}
\sum_{j=1}^{|\yans|}
\phi\!\left(\mathrm{BALD}_j\right).
$
The resulting estimator depends on the dropout rate, the layers in which
dropout is enabled, the number of stochastic passes, and the choice of
$\phi$; these implementation details are provided in
Appendix~\ref{app:bald}.

% Since our goal is to estimate confidence in the correctness of the final answer, we aggregate token-level BALD scores over all answer tokens. Additionally, the BALD score is an uncertainty measure and is not naturally bounded in $[0,1]$. To enable comparison with other confidence estimators, we convert it into a confidence score via a monotonic transformation:
% \begin{equation}\label{eq:c_bald}
%     c_{\text{BALD}} = \frac{1}{|\yans|}\sum_{j=1}^{|\yans|}\mathrm{Normalize}(-\mathrm{BALD}_j),
% \end{equation}
% where higher values indicate greater confidence in the predicted answer. In Appendix~\ref{app:bald}, we include further discussion on normalization challenges and implementation details.

% Both $c_{\mathrm{BALD}}$ and $\ptrue$ require additional forward passes, in contrast to the single-pass estimators in Section~\ref{sec:single-pass}. Table~\ref{tab:estimators} summarizes these estimators. Together, these methods span different ways of extracting confidence from model output probabilities: direct aggregation over generated tokens, self-verification via re-prompting, and uncertainty estimation via MC Dropout. This provides a unified framework for systematically studying token-probability-based confidence estimation.

Like $\ptrue$, $c_{\mathrm{BALD}}$ requires additional forward passes.
Together, these methods allow us to compare
confidence derived directly from the original generation, self-verification,
and stochastic predictive disagreement.

\section{Confidence Calibration}\label{sec:calibration}
Calibration aims to align confidence estimates with empirical correctness~\citep{guo2017calibration}. A confidence estimator $c(\bm{x}, \hat{\bm{y}})$ is \textit{perfectly calibrated} if
$
\mathbb{P}\!\left(\mathbb{I}(\hat{\bm{y}}^{\mathrm{ans}}, \bm{y}^*) = 1 \mid c(\bm{x}, \hat{\bm{y}})=p\right)=p
$
for all $p \in [0,1]$. Since $c(\bm{x}, \hat{\bm{y}})$ is typically continuous, this condition is evaluated empirically by partitioning predictions into confidence bins.

\textbf{Metrics:}
Qualitatively, reliability diagrams plot empirical accuracy against mean confidence across bins; perfect calibration lies on the diagonal~\citep{niculescu2005predicting}. Quantitatively, Expected Calibration Error (ECE) and Maximum Calibration Error (MCE)~\citep{naeini2015obtaining} measure the average and worst-case discrepancies across these bins. We defer results with Brier scores~\citep{glenn1950verification} to Appendix~\ref{app:additional_result}.

\textbf{Post-hoc calibration} learns a mapping $f$ on a held-out calibration set so that
$f(c(\bm{x},\hat{\bm{y}}))$ better approximates the probability of answer
correctness. Following prior work~\citep{guo2017calibration}, we evaluate two simple and widely used approaches, isotonic regression and Platt scaling, and study their behavior across calibration-set sizes, dataset difficulty, and cross-dataset transfer. \textit{Isotonic regression}~\citep{zadrozny2002transforming} learns a flexible, non-parametric monotonic mapping. Because it requires only confidence scores and correctness labels, it can fit non-linear calibration curves and is applied to all estimators in our study. \textit{Platt scaling}~\citep{platt1999probabilistic} applies a parametric global adjustment to logits. Since correctness labels are available only for final answers, 
% we apply it to answer-token estimators via temperature scaling and to $\ptrue$ via a sigmoid map over the logit margin.
we apply temperature scaling for answer-token estimators and Platt scaling for the binary verification score.

% In LLMs, confidence is constructed from token probabilities rather than native classification scores, raising the question of whether standard calibration methods remain effective across diverse confidence estimators~\citep{jiang2021can, chen2023close, zhu2023calibration}. 
\section{Experiments}\label{sec:exp}
In this section, we evaluate confidence estimators and analyze the factors driving their calibration performance. 
% We also study post-hoc calibration, showing that it can recover useful confidence signals from poorly calibrated estimators and that calibration data size depend on dataset difficulty.
We further study post-hoc calibration, including its data efficiency, sensitivity to dataset difficulty, and transfer across datasets.

\subsection{Experiment Setup}\label{sec:exp-setup}
\noindent\textbf{Model Selection:}
We focus on open-source language models, including 
Llama-3.1-8B-Instruct, 
Llama-3.2-3B-Instruct~\citep{dubey2024llama},
Qwen3-8B-Thinking, 
Qwen3-4B-Instruct~\citep{yang2025qwen3},
and Deepseek-R1-distill-llama-8b~\citep{guo2025deepseek}.
This collection spans a range of model sizes, architectures, and includes
instruction-tuned, distilled, and reasoning models.

\noindent\textbf{Datasets:}
We evaluate on GSM8K~\citep{cobbe2021training}, GSMHard~\citep{gao2022pal}, and SVAMP~\citep{patel2021nlp}, three mathematical question answering benchmarks with varying difficulty and over two thousand evaluation examples in total. These tasks elicit structured outputs with intermediate reasoning followed by a final numeric answer~\citep{wei2022chain}, enabling verifiable answer extraction for evaluating correctness.

\noindent\textbf{Estimation:}
We use each model's recommended chat template and report the decoding configuration in Table~\ref{tab:inference_settings}. For $p(\text{True})$, we follow the re-prompting template~\citep{kadavath2022language} and identify the token indices corresponding to \texttt{True} and \texttt{False} for each model. 
For MC Dropout estimation, we select the dropout rate separately for each model through grid search and apply it to all supported dropout modules during stochastic forward passes. Exploring layer-specific dropout configurations is left for future work.
% For MC Dropout-based estimation, we select the dropout rate via grid search. Dropout is enabled for all model parameters; exploring more fine-grained dropout configurations is left for future work.

\subsection{Confidence Estimation}

\begin{table*}[t]
\centering
\renewcommand{\arraystretch}{0.8}
\setlength{\tabcolsep}{3.7pt}
\fontsize{8pt}{9.3pt}\selectfont
\begin{tabular}{llc | cc cc cc cc | cc cc}
\toprule
\multirow{2}{*}{Model} & \multirow{2}{*}{Dataset} & \multirow{2}{*}{Acc.} 
& \multicolumn{2}{c}{\textsc{Seq Avg}} 
& \multicolumn{2}{c}{\textsc{Seq Joint}} 
& \multicolumn{2}{c}{\textsc{Ans Avg}} 
& \multicolumn{2}{c|}{\textsc{Ans Joint}}
& \multicolumn{2}{c}{$\ptrue$} 
& \multicolumn{2}{c}{$c_{\mathrm{BALD}}$} \\
\cmidrule(lr){4-5} \cmidrule(lr){6-7} \cmidrule(lr){8-9} \cmidrule(lr){10-11}
\cmidrule(lr){12-13} \cmidrule(lr){14-15}
 &  &  
& ECE & MCE 
& ECE & MCE 
& ECE & MCE 
& ECE & MCE 
& ECE & MCE 
& ECE & MCE \\
\midrule

\multirow{3}{*}{Llama-3.1-8B}
& GSM8K   & 0.824 & \textbf{0.053} & 0.481 & 0.117 & 0.195 & 0.166 & 0.529 & 0.165 & 0.469 & 0.097 & 0.625 & 0.149 & 0.667 \\
& GSMHard & 0.350 & 0.457 & 0.600 & 0.320 & 0.414 & 0.622 & 0.743 & 0.615 & 0.753 & 0.333 & 0.627 & \textbf{0.152} & 0.424 \\
& SVAMP   & 0.827 & \textbf{0.057} & 0.309 & 0.170 & 0.332 & 0.169 & 0.797 & 0.171 & 0.765 & 0.099 & 0.450 & 0.095 & 0.633 \\

\midrule
\multirow{3}{*}{Llama-3.2-3B}
& GSM8K   & 0.775 & 0.124 & 0.587 & 0.092 & 0.364 & 0.218 & 0.635 & 0.218 & 0.623 & \textbf{0.055} & 0.304 & 0.108 & 0.389 \\
& GSMHard & 0.258 & 0.636 & 0.773 & 0.597 & 0.676 & 0.725 & 0.857 & 0.722 & 0.768 & 0.251 & 0.406 & \textbf{0.078} & 0.202 \\
& SVAMP   & 0.820 & 0.068 & 0.783 & \textbf{0.035} & 0.676 & 0.175 & 0.840 & 0.175 & 0.837 & 0.104 & 0.423 & 0.097 & 0.610 \\

\midrule
\multirow{3}{*}{DeepSeek-R1-8B}
& GSM8K   & 0.650 & 0.253 & 0.384 & 0.219 & 0.476 & 0.304 & 0.361 & 0.302 & 0.356 & 0.315 & 0.484 & \textbf{0.168} & 0.732 \\
& GSMHard & 0.320 & 0.588 & 0.646 & 0.556 & 0.611 & 0.637 & 0.674 & 0.631 & 0.690 & 0.634 & 0.673 & \textbf{0.102} & 0.606 \\
& SVAMP   & 0.728 & 0.164 & 0.784 & \textbf{0.127} & 0.525 & 0.244 & 0.617 & 0.242 & 0.606 & 0.241 & 0.536 & 0.174 & 0.979 \\

\midrule
\multirow{3}{*}{Qwen3-8B-Thinking}
& GSM8K   & 0.958 & \textbf{0.006} & 0.070 & 0.018 & 0.027 & 0.042 & 0.042 & 0.042 & 0.042 & 0.035 & 0.829 & 0.043 & 0.545 \\
& GSMHard & 0.810 & 0.144 & 0.781 & 0.128 & 0.793 & 0.190 & 0.190 & 0.190 & 0.190 & 0.183 & 0.373 & \textbf{0.028} & 0.365 \\
& SVAMP   & 0.972 & \textbf{0.017} & 0.017 & 0.033 & 0.034 & 0.028 & 0.028 & 0.028 & 0.028 & 0.020 & 0.348 & 0.026 & 0.133 \\

\midrule
\multirow{3}{*}{Qwen3-4B-Instruct}
& GSM8K   & 0.938 & 0.035 & 0.478 & \textbf{0.028} & 0.790 & 0.062 & 0.062 & 0.062 & 0.062 & 0.062 & 0.652 & 0.061 & 0.650 \\
& GSMHard & 0.701 & 0.266 & 0.724 & 0.256 & 0.787 & 0.298 & 0.299 & 0.298 & 0.299 & 0.285 & 0.985 & \textbf{0.188} & 0.451 \\
& SVAMP   & 0.943 & 0.032 & 0.366 & \textbf{0.025} & 0.267 & 0.057 & 0.057 & 0.057 & 0.057 & 0.058 & 0.999 & 0.061 & 0.313 \\

\bottomrule
\end{tabular}
\caption{
\textbf{Confidence estimation results using token probabilities.} 
We report accuracy (Acc.) and calibration errors (ECE, MCE; lower is better).
For single-pass methods, sequence-based estimators achieve lower ECE than answer-based estimators across all model--dataset pairs.
For multi-pass methods, MC Dropout-based $c_{\mathrm{BALD}}$ often achieves lower ECE than $\ptrue$, particularly on GSMHard.
\textbf{Bold} indicates the lowest ECE within each row.
}
\label{tab:result_estimation}
\end{table*}

We investigate the effectiveness of token probability in confidence estimation.
Table~\ref{tab:result_estimation} summarizes results for single-pass and multi-pass confidence estimators together with model accuracy on each dataset. Full results, including Brier scores and bootstrap confidence intervals, are reported in  Table~\ref{tab:result_estimation_full} in the appendix.
% \footnote{DeepSeek-R1's degraded performance might come from a mismatch between its distillation format and the evaluation setup. We discuss this in Appendix~\ref{app:implementation}.} 
% We highlight several key findings.

\begin{findingbox}
\textbf{Finding 1 (Trajectory-level Confidence):} 
% Answer-token probabilities are often saturated, while 
Calibrated confidence arises from aggregating weak but consistent probability differences across the reasoning trajectory.
\end{findingbox}
Our analysis of single-pass estimators considers two design choices: the subset of generated tokens used for estimation and the method used to aggregate their probabilities. Across all model--dataset pairs, sequence-based estimators achieve lower ECE than answer-only estimators, although the confidence intervals overlap in some settings.

% Answer-token estimators are particularly affected by overconfidence, consistent with prior findings on LLM confidence~\citep{chhikara2025mind}. 
% Figure~\ref{fig:main_analysis}(a) illustrates this behavior: answer-token probabilities are concentrated near $1.0$ for both correct and incorrect outputs, providing limited separation between the two groups.

\begin{figure}[t]
    % \captionsetup[subfigure]{labelformat=empty, labelsep=period}
    \centering
    {\includegraphics[width=\linewidth]{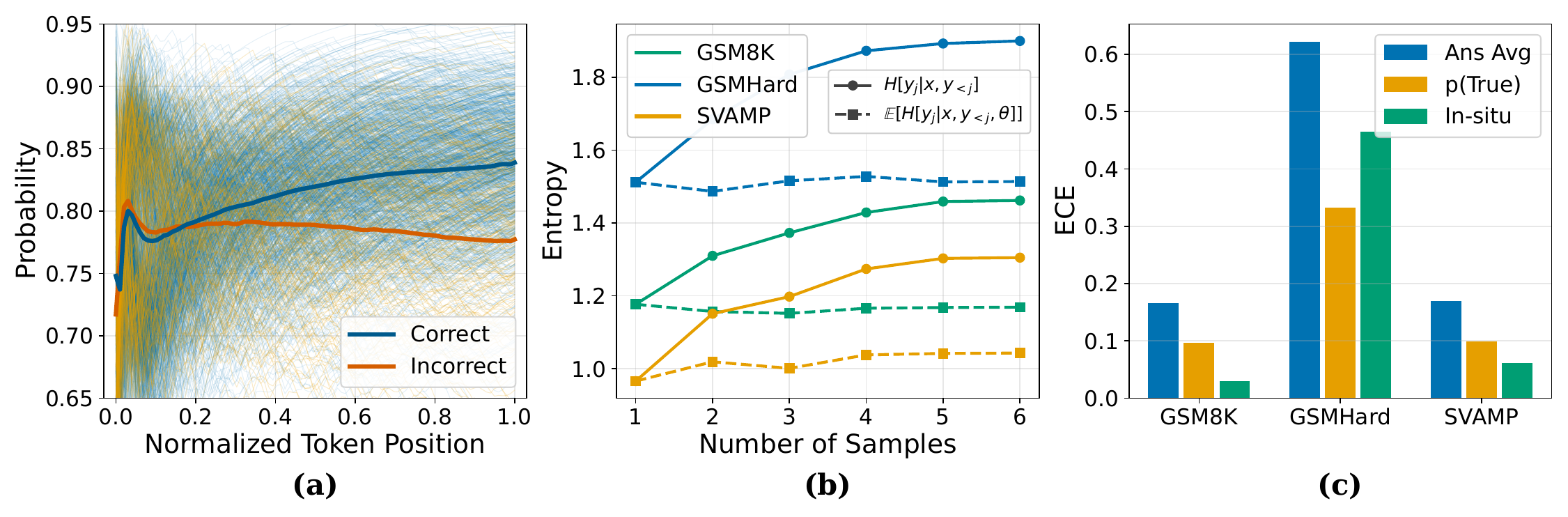}}
    \vspace{-0.5cm}
    \caption{\textbf{Confidence estimation analysis.}
    % (a) \textbf{Density of answer-token probabilities for correct and incorrect outputs.} 
    % Probabilities are heavily concentrated near 1.0, providing limited separation between correct and incorrect answers.
    (a) \textbf{Moving average of token probabilities across the generated sequence.} 
    Confidence-relevant information arises from the accumulation of small token-level probability differences across the reasoning trajectory.
    (b) \textbf{BALD decomposition under MC Dropout.} 
    Predictive entropy increases and stabilizes after roughly five samples, indicating that MC Dropout primarily captures distributional uncertainty across stochastic inferences.
    (c) \textbf{In-situ verification.} 
    In-situ $\ptrue$ achieves calibration comparable to standard $\ptrue$ while avoiding full re-prompting. Results are from Llama-3.1-8B, with (a) evaluated on GSM8K.
    % \leila{which dataset was used for these results?\fixed}
    }
    
    \label{fig:main_analysis}
    \vspace{-0.2cm}
\end{figure}

% In contrast, leveraging token probabilities over the entire generated sequence substantially improves calibration. In many cases, sequence-level estimators produce lower ECE, indicating confidence scores that better reflect output correctness, and can even outperform more expensive multi-pass estimators.
Using probabilities from the full generated sequence substantially improves calibration. In several settings, sequence-based estimators also achieve lower ECE than the more computationally expensive multi-pass estimators.
To understand this, we examine the trajectory of token probabilities throughout generation. In Figure~\ref{fig:main_analysis}(a), we plot the moving average of token probabilities as a function of normalized token position. This can be viewed as a \textsc{Seq Avg} estimator computed with an increasing window size. The curves initially overlap because many responses share similar opening phrases. As generation progresses, however, the curves separate: average probabilities tend to increase for responses with correct final answers and decrease for those with incorrect answers.
% For outputs that lead to correct answers, the moving average increases, suggesting that the model remains on a coherent reasoning path and continues to assign high probability to subsequent tokens. For incorrect outputs, the moving average gradually decreases, suggesting that the model drifts away from a reliable solution path and assigns lower probability to subsequent tokens. 

% Although individual token probabilities can be saturated, this growing separation in Figure~\ref{fig:main_analysis}(b) suggests that confidence signals arise from the accumulation of small token-level probability differences across the reasoning trajectory. While such differences are difficult to distinguish at the individual-token level, they become more separable when aggregated over the full generated sequence.
This pattern suggests that the useful signal does not necessarily arise from any single token. Rather, small probability differences accumulate across the reasoning trajectory and become more apparent after sequence-level aggregation.

Within each token subset, 
% joint aggregation often outperforms average aggregation, especially on more challenging datasets such as GSMHard. Since length-normalized joint probability and average probability correspond to geometric and arithmetic means, respectively, this suggests that confidence estimation benefits from penalizing low-probability tokens more strongly.
% 
% That is, rare low-probability tokens along the reasoning trajectory can be informative indicators of potential errors, and geometric aggregation is more sensitive to such steps.
length-normalized joint probability often outperforms arithmetic averaging, particularly on GSMHard. Because the joint probability is more sensitive to low-probability tokens, this result suggests that occasional low-probability steps can provide useful evidence of answer incorrectness.

\noindent\textbf{Actionable takeaway.} 
Full-trajectory aggregation provides a strong single-pass default, suggesting that useful confidence signals are distributed across the reasoning trajectory.
Because global aggregation obscures where these signals arise, local-window methods offer a natural extension. Recent work similarly exploits localized confidence patterns, but primarily to improve test-time reasoning efficiency rather than confidence calibration~\citep{fu2026deep}. Extending such local-window approaches to produce calibrated estimates of correctness is a promising direction for future work.

\begin{findingbox}
\textbf{Finding 2 (Distributional Uncertainty):}
MC Dropout captures uncertainty from variation in predictive distributions, but requires dropout-rate selection for each model--dataset pair and multiple stochastic passes.
\end{findingbox}
We next compare two multi-pass confidence estimators: $\ptrue$, which re-prompts the model to verify its generated answer, and $c_{\mathrm{BALD}}$, which measures dropout-induced variation across token predictive distributions. As shown in Table~\ref{tab:result_estimation}, $c_{\mathrm{BALD}}$ often achieves lower calibration error than $\ptrue$, with its advantage being
particularly pronounced on GSMHard. However, these gains come with substantial computational and tuning costs.

Both methods introduce additional computational overhead compared to single-pass estimators. For $\ptrue$, the extra cost is roughly proportional to the length of the re-encoded question--answer pair. For $c_{\mathrm{BALD}}$, inference cost grows approximately linearly with the number of stochastic passes $T$. Moreover, its performance is sensitive to the dropout rate, which we select separately for each model--dataset pair. The search space, criteria, and the selected dropout rates are reported in Appendix~\ref{app:dropout}.

To examine the trade-off in the number of stochastic passes, Figure~\ref{fig:main_analysis}(b) plots the two terms of Equation~\eqref{eq:bald} as $T$ increases. The expected conditional entropy, $\mathbb{E}_{\theta}[H(y_j \mid x, y_{<j}, \theta)]$, remains relatively stable, indicating that individual dropout-induced forward passes produce similarly sharp distributions. In contrast, the predictive entropy, $H(y_j \mid x, y_{<j})$, increases with $T$ and stabilizes after roughly five samples, resulting in about a five-fold overhead relative to a single forward pass for a reliable estimation.

\noindent\textbf{Actionable takeaway.}
MC Dropout is most appropriate when its calibration gains justify substantial tuning and inference costs. Recent approaches reduce repeated inference~\citep{gao2025flue}, suggesting a promising direction toward more efficient confidence estimation.

% Importantly, this variation does not necessarily correspond to changes in the decoded answer. In our setting, the top predicted token remains unchanged across stochastic forward passes; instead, dropout mainly changes how probability mass is distributed across tokens. Thus, $c_{\mathrm{BALD}}$ improves calibration by capturing variation in predictive distributions of the same generated answer, rather than relying on output-level inconsistency as in consistency-based methods~\citep{xiong2023can, wang2022self, lyu2025calibrating}. This provides a confidence signal unavailable from a single deterministic pass, though at substantially higher inference cost.

\begin{findingbox}
\textbf{Finding 3 (In-situ Self-verification):}
In-situ self-verification achieves calibration comparable to standard $\ptrue$ with substantially lower token-processing overhead.
\end{findingbox}
The performance of $\ptrue$ raises a natural question: why is it often better calibrated than \textsc{Ans Avg}, despite both relying on only a few token probabilities? One possible explanation is that $\ptrue$ reframes confidence estimation as binary self-verification, a setting in which language models have been shown to provide informative confidence signals~\citep{kadavath2022language}.

% Motivated by this interpretation, we transfer the self-verification framing of $\ptrue$ to the original generation trajectory. Specifically, we evaluate an \textit{in-situ} variant that appends the binary verification query directly to the original response, rather than re-prompting the model from scratch, and computes confidence using the $\ptrue$ estimator in Table~\ref{tab:estimators}. This performs verification \textit{in situ}: the verifier remains conditioned on the same inference trajectory that produced the final answer, including the inference configuration, system prompt, one-shot template, question, and generated response. A comparison is provided in Figure~\ref{fig:prompt_comparison} in the Appendix.

Motivated by this observation, we evaluate an \textit{in-situ} variant that appends the verification query to the original response rather than re-prompting the model from scratch. The resulting verifier remains conditioned on the original prompt and generated trajectory, including the system prompt, demonstration, question, reasoning, and final answer. The prompt formats are compared in Figure~\ref{fig:prompt_comparison}.
% As shown in Figure~\ref{fig:main_analysis}(d), the in-situ variant achieves ECE comparable to standard $\ptrue$ while avoiding re-encoding the full question--answer pair and reducing token-processing overhead by 88\% on average across all models and datasets. Additional results are provided in Table~\ref{tab:insitu_ptrue}. This is particularly useful for practical applications involving long outputs~\citep{wu2024longgenbench, bai2024longwriter}, where the computational overhead of standard re-prompting grows with response length and can become substantial.

As shown in Figure~\ref{fig:main_analysis}(c), in-situ verification achieves ECE comparable to standard $\ptrue$. By reusing the original generation, it avoids reprocessing the full question--response sequence and reduces token-processing overhead by 88\% on average across all settings. Full results are reported in Table~\ref{tab:insitu_ptrue}. This saving is particularly relevant for long outputs~\citep{bai2024longwriter}.

\noindent\textbf{Actionable takeaway.}
When the original generation state can be retained, in-situ verification provides a substantially cheaper alternative to standard $\ptrue$, especially for long outputs.

\begin{table*}[t]
\centering
\renewcommand{\arraystretch}{1}
\renewcommand{\tabcolsep}{3.2pt}

\scriptsize
\resizebox{\textwidth}{!}{
\begin{tabular}{ll | ccc | ccc cc}
\toprule
\multirow{2}{*}{Model} 
& \multirow{2}{*}{Dataset} 
& \multicolumn{3}{c|}{Isotonic Regression} 
& \multicolumn{5}{c}{Platt Scaling} \\
\cmidrule(lr){3-5} \cmidrule(lr){6-10}
& 
& \textsc{Ans Avg} 
& \textsc{Ans Joint} 
& $\ptrue$ 
& $t$ 
& \textsc{Ans Avg} 
& \textsc{Ans Joint} 
& $(a,b)$ 
& $\ptrue$ \\
\midrule

\multirow{3}{*}{Llama-3.1-8B}
& GSM8K   
& \textbf{0.056 (+66\%)} & \textbf{0.054 (+67\%)} & \textbf{0.058 (+40\%)} 
& 1.82 & 0.080 (+52\%) & 0.080 (+51\%) & (0.53, 0.48) & 0.059 (+39\%) \\
& GSMHard 
& \textbf{0.064 (+90\%)} & 0.065 (+89\%) & \textbf{0.057 (+83\%)} 
& 2.37 & 0.091 (+85\%) & \textbf{0.045 (+93\%)} & (0.57, -1.74) & 0.087 (+74\%) \\
& SVAMP   
& \textbf{0.020 (+88\%)} & \textbf{0.022 (+87\%)} & 0.045 (+54\%) 
& 1.91 & 0.084 (+50\%) & 0.083 (+52\%) & (0.54, 0.43) & \textbf{0.039 (+60\%)} \\

\midrule

\multirow{3}{*}{Llama-3.2-3B}
& GSM8K   
& \textbf{0.074 (+66\%)} & \textbf{0.075 (+65\%)} & 0.076 (-39\%) 
& 1.74 & 0.104 (+52\%) & 0.103 (+53\%) & (0.65, 1.07) & \textbf{0.044 (+20\%)} \\
& GSMHard 
& \textbf{0.027 (+96\%)} & \textbf{0.028 (+96\%)} & \textbf{0.031 (+88\%)} 
& 2.27 & 0.105 (+85\%) & 0.057 (+92\%) & (0.79, -1.36) & 0.047 (+81\%) \\
& SVAMP   
& \textbf{0.036 (+79\%)} & \textbf{0.036 (+80\%)} & 0.055 (+48\%) 
& 1.79 & 0.044 (+75\%) & 0.041 (+76\%) & (0.61, 0.80) & \textbf{0.029 (+72\%)} \\

\midrule

\multirow{3}{*}{DeepSeek-R1-8B}
& GSM8K   
& \textbf{0.012 (+96\%)} & \textbf{0.013 (+96\%)} & 0.017 (+95\%) 
& 1.68 & 0.071 (+77\%) & 0.070 (+77\%) & (0.054, 0.46) & \textbf{0.016 (+95\%)} \\
& GSMHard 
& \textbf{0.068 (+89\%)} & \textbf{0.071 (+89\%)} & 0.066 (+90\%) 
& 2.15 & 0.100 (+84\%) & 0.082 (+87\%) & (0.065, -1.29) & \textbf{0.065 (+90\%)} \\
& SVAMP   
& \textbf{0.016 (+93\%)} & \textbf{0.024 (+90\%)} & 0.034 (+86\%) 
& 1.67 & 0.061 (+75\%) & 0.066 (+73\%) & (0.022, 0.83) & \textbf{0.013 (+95\%)} \\

\midrule

\multirow{3}{*}{Qwen3-8B-Thinking}
& GSM8K   
& \textbf{0.015 (+63\%)} & \textbf{0.015 (+63\%)} & \textbf{0.015 (+58\%)} 
& 3.19 & 0.030 (+29\%) & 0.030 (+28\%) & (0.39, 1.54) & 0.016 (+54\%) \\
& GSMHard 
& \textbf{0.034 (+82\%)} & \textbf{0.033 (+83\%)} & \textbf{0.049 (+73\%)} 
& 3.62 & 0.114 (+40\%) & 0.114 (+40\%) & (0.52, -1.38) & 0.078 (+57\%) \\
& SVAMP   
& \textbf{0.003 (+91\%)} & \textbf{0.003 (+91\%)} & 0.007 (+66\%) 
& 3.22 & 0.022 (+20\%) & 0.022 (+21\%) & (0.46, 1.22) & \textbf{0.004 (+79\%)} \\

\midrule

\multirow{3}{*}{Qwen3-4B-Instruct}
& GSM8K   
& \textbf{0.019 (+69\%)} & \textbf{0.019 (+69\%)} & 0.020 (+68\%) 
& 3.90 & 0.044 (+29\%) & 0.044 (+29\%) & (0.18, 1.70) & \textbf{0.019 (+70\%)} \\
& GSMHard 
& \textbf{0.077 (+74\%)} & \textbf{0.077 (+74\%)} & \textbf{0.091 (+68\%)} 
& 4.48 & 0.131 (+56\%) & 0.144 (+52\%) & (0.11, -0.18) & 0.092 (+68\%) \\
& SVAMP   
& \textbf{0.035 (+38\%)} & \textbf{0.035 (+38\%)} & 0.032 (+44\%) 
& 3.88 & 0.043 (+24\%) & 0.044 (+22\%) & (-0.30, 6.02) & \textbf{0.032 (+45\%)} \\

\bottomrule
\end{tabular}
}
\caption{
\textbf{Post-hoc calibration results.}
We report calibrated ECE for \textsc{Ans Avg}, \textsc{Ans Joint}, and $\ptrue$ after isotonic regression and Platt scaling, with relative improvements over uncalibrated estimators shown in parentheses.
For Platt scaling, we also report the learned temperature $t$ for answer-based estimators and learned parameters $(a,b)$ for $\ptrue$.
For each estimator and dataset, \textbf{bold} indicates the lower ECE between the two calibration methods.
}
\label{tab:result_calibration}
\end{table*}

\subsection{Confidence Calibration}

We next evaluate post-hoc calibration across confidence estimators. We consider isotonic regression and Platt scaling. Table~\ref{tab:result_calibration} reports the calibrated ECE for \textsc{Ans Avg}, \textsc{Ans Joint}, and $\ptrue$, with relative improvements over the corresponding uncalibrated estimators shown in parentheses. Complete results with confidence intervals are deferred to Table~\ref{tab:result_calibration_full} in the appendix.

\begin{figure}[t]
    % \captionsetup[subfigure]{labelformat=empty, labelsep=period}
    \centering
    {\includegraphics[width=\linewidth]{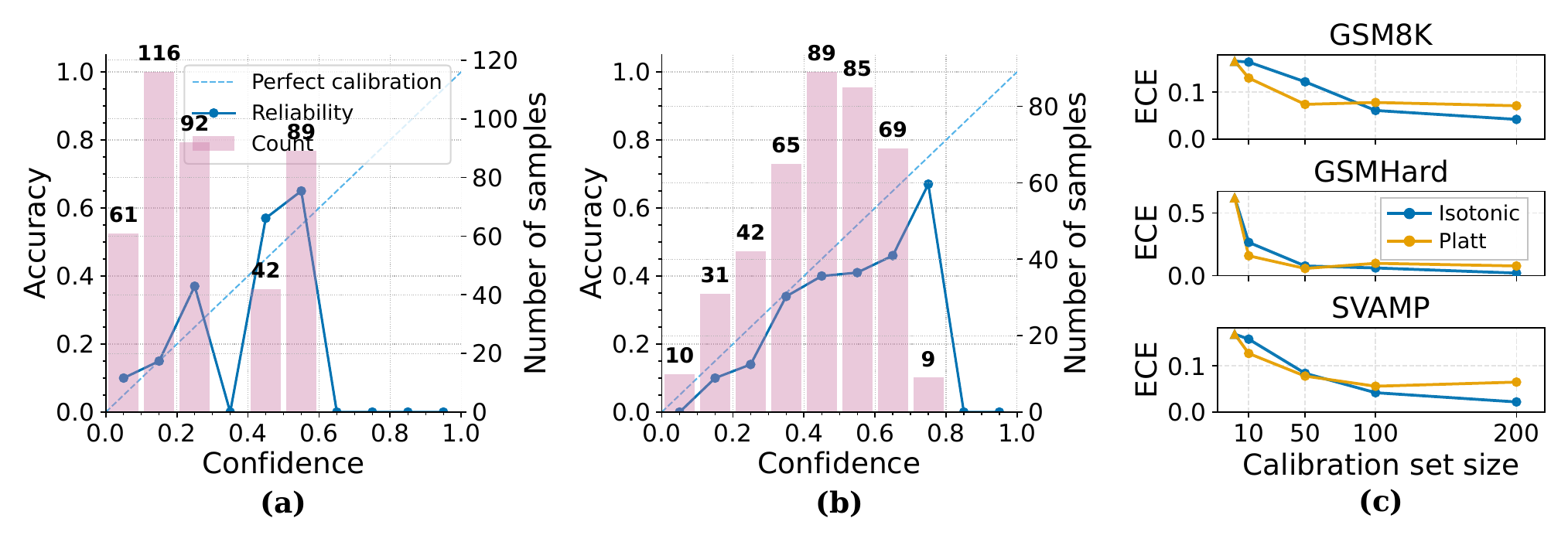}}
    \vspace{-0.5cm}
    \caption{
    \textbf{Post-hoc confidence calibration analysis.}
    Reliability diagrams for Ans Avg: both 
    % (a) \textbf{before calibration}, 
    (a) \textbf{isotonic regression} and (b) \textbf{Platt scaling} improve calibration, with isotonic regression fitting a flexible monotonic mapping and Platt scaling applying a global parametric correction.
    (c) \textbf{Effect of calibration set size on Ans Avg.} Calibration error decreases quickly with calibration set size, with 50 examples already yielding substantial improvements across datasets.
    Results are from Llama-3.1-8B, with (a) and (b) evaluated on GSMHard.
    }

    \label{fig:calibration_analysis}
    \vspace{-0.2cm}
\end{figure}

Both methods generally reduce calibration error, including the strongly overconfident answer-token estimators. Isotonic regression often produces the largest reductions. As illustrated in Figure~\ref{fig:calibration_analysis}, its flexible monotonic mapping can correct nonlinear reliability curves that cannot be captured by the global logit rescaling of temperature scaling. These results show that high raw ECE does not necessarily preclude effective post-hoc correction when labeled calibration data are available.

% \begin{findingbox}
% \textbf{Finding 4 (Recovering Poor Estimators):} Even poorly calibrated confidence estimators can be substantially improved by post-hoc calibration, suggesting their raw scores still contain useful signals about correctness.
% \end{findingbox}

\begin{findingbox}
\textbf{Finding 4 (Difficulty-dependent Calibration and Transfer):}
Calibration can be more data-efficient on harder datasets under random sampling, but learned mappings transfer asymmetrically across datasets and models.
\end{findingbox}

We first study how calibration-set size affects performance. Prior work often characterizes calibration-data requirements in terms of the calibration method~\citep{guo2017calibration} or label noise~\citep{zhao2020role}. We complement these analyses by examining the role of dataset difficulty.

% Both isotonic regression and Platt scaling improve calibration across models, datasets, and estimators. As shown in Table~\ref{tab:result_calibration}, isotonic regression often produces the largest reductions. The gains are particularly substantial for \textsc{Ans Avg} and \textsc{Ans Joint}, whose raw scores exhibit the highest calibration errors.

% Figure~\ref{fig:calibration_analysis} illustrates why isotonic regression can be particularly effective. For \textsc{Ans Avg}, it learns a flexible monotonic mapping that closely corrects the reliability curve, whereas temperature scaling applies only a global rescaling of the token logits. This added flexibility allows isotonic regression to better correct strongly nonlinear miscalibration.

% \noindent\textbf{Actionable takeaway.}
% Raw ECE alone should not be used to discard a confidence estimator. When held-out correctness labels are available, post-hoc calibration can substantially improve its reliability, with isotonic regression being particularly effective for strongly nonlinear miscalibration.

% \begin{findingbox}
% \textbf{Finding 5 (Dataset-dependent Calibration Efficiency):}
% Calibration data efficiency depends on dataset difficulty. Under random sampling, easier datasets may require larger calibration sets for post-hoc calibration to become effective.
% \end{findingbox}
% Lastly, we study how calibration set size affects post-hoc calibration performance. Prior work often discusses calibration data requirements in terms of the calibration method~\cite{guo2017calibration} and label noise~\citep{zhao2020role}. We complement by showing that calibration data efficiency also depends on dataset difficulty.

Figure~\ref{fig:calibration_analysis}(c) plots the ECE of \textsc{Ans Avg} after isotonic regression and Platt scaling as the calibration set size varies. Calibration examples are randomly sampled from the calibration split. Overall, calibration error decreases rapidly with more calibration data, with 50 examples already yielding substantial improvements across datasets. However, GSMHard benefits from calibration with fewer examples, whereas GSM8K and SVAMP require larger calibration sets for stable improvement.

This is because the model is more accurate on GSM8K and SVAMP: under random sampling, small calibration sets contain many high-confidence correct examples, making overconfidence harder to estimate. Harder datasets, however, expose high-confidence errors more frequently, allowing effective calibration with fewer examples. This suggests an important direction for future work: calibration-set construction need not rely on uniform random sampling, and weighted sampling strategies that emphasize incorrect validation examples may improve calibration data efficiency.

\begin{table}[t]
\centering
\scriptsize
\setlength{\tabcolsep}{8pt}
\begin{tabular}{lrr}
\toprule
Calibration Source $\rightarrow$ Evaluation Target & Isotonic & Platt \\
\midrule
    Llama: GSM8K $\rightarrow$ Llama: GSMHard & $\bm{-.209}$ & $\bm{-.083}$ \\
    Llama: GSMHard $\rightarrow$ Llama: GSM8K & $+.189$ & $+.213$ \\
    Qwen: GSM8K $\rightarrow$ Qwen: GSMHard & $\bm{-.034}$ & $\bm{-.054}$ \\
    Qwen: GSMHard $\rightarrow$ Qwen: GSM8K & $+.242$ & $-.011$ \\
\midrule
    Llama: GSM8K  $\rightarrow$ Qwen:GSM8K & $-.003$ & $-.002$ \\
    Qwen: GSM8K  $\rightarrow$ Llama:GSM8K & $+.458$ & $+.658$ \\
    Llama: GSMHard $\rightarrow$ Qwen:GSMHard & $\bm{-.185}$ & $\bm{-.003}$ \\
    Qwen: GSMHard $\rightarrow$ Llama:GSMHard & $\bm{-.527}$ & $\bm{-.318}$ \\
\bottomrule
\end{tabular}
\caption{
\textbf{Off-domain calibration transfer for \textsc{Ans Avg}.}
Each entry reports $\Delta\mathrm{ECE}$ relative to the uncalibrated target;
negative values indicate improvement. Full results are included in Table~\ref{tab:calibration_transfer_full} in the appendix.
}
\label{tab:calibration_transfer_summary}

\end{table}
We further examine calibration transfer across datasets and models. For \textsc{Ans Avg}, we fit calibrators using Llama-3.1-8B-Instruct and Qwen3-4B-Instruct on either GSM8K or GSMHard, then apply each mapping unchanged to a different dataset or model. Table~\ref{tab:calibration_transfer_summary} reports the change in target ECE relative to the uncalibrated estimator.

Cross-dataset transfer is strongly asymmetric: mappings learned on GSM8K consistently improve calibration on GSMHard, whereas the reverse direction generally worsens it. Cross-model transfer is similarly target-dependent. Both directions improve calibration on GSMHard, while transfer on GSM8K is asymmetric and can severely degrade calibration. These results suggest that calibration mappings depend on both model-specific score distributions and dataset difficulty, limiting their reliability under transfer.

\noindent\textbf{Actionable takeaway.}
Calibration data should match the target model and difficulty level whenever possible. When the model is known to be overconfident and calibration labels are limited, difficulty-aware sampling may improve efficiency by ensuring sufficient coverage of model errors. 
% Off-domain calibration mappings should be validated on the target setting before deployment, as transfer can substantially worsen calibration.
\section{Conclusion}

In this work, we study token-probability-based confidence estimation and calibration for mathematical question answering. We compare single-pass and multi-pass confidence estimators, then
evaluate post-hoc calibration. Our findings show that isotonic regression and Platt scaling generally reduce in-domain calibration error, although their data efficiency and transferability depend on the model and dataset. Overall, our results show that token selection, aggregation, inference cost, and calibration data all affect the quality and practicality of token-probability-based confidence estimates.

% % \input{sec/x_ethics}
\section*{Limitations}
Our study evaluates open-weight LLMs for which token probabilities and dropout controls are accessible, limiting direct evaluation of proprietary models. Our study focuses on mathematical question answering tasks with verifiable final answers, but open-ended dialogue requires handling more ambiguous and task-dependent notions of correctness. We consider two classical post-hoc calibration families, isotonic regression and parametric scaling. Although they reveal important challenges in calibration efficiency and transfer, more expressive calibration methods may behave differently.

The MC Dropout results also depend on several design choices: conditioning on a fixed reasoning trajectory, measuring uncertainty only at answer-token positions, and selecting the dropout rate, enabled layers, number of stochastic passes, and normalization function. This formulation reduces computation but does not capture uncertainty over alternative reasoning trajectories. Correctness evaluation relies on regex-based answer extraction using a fixed delimiter; malformed or nonconforming outputs may therefore introduce extraction errors. Finally, our experiments use controlled prompts, whereas real deployments may involve long contexts, multi-turn histories, and external evidence, whose effects on token-probability-based confidence estimation remain open.

\bibliography{reference}

\appendix
\appendix

\section{Summary of the Supplementary Material}

The supplementary material is organized as follows.
In Appendix~\ref{app:bald}, we provide additional details on how the BALD uncertainty score is converted into the confidence estimator $c_{\mathrm{BALD}}$, including the normalization strategies we considered and the implementation choices used in our main experiments.
In Appendix~\ref{app:dropout}, we discuss the dataset, the selection criteria, and present the dropout sweep results for all model--dataset pairs.
In Appendix~\ref{app:implementation}, we describe additional experimental details, including dataset splits, answer extraction, prompting, calibration metric computation, bin-count analysis, and model-specific inference settings.
Finally, Appendix~\ref{app:additional_result} presents additional experimental results, including full confidence-estimation and post-hoc calibration results with bootstrap confidence intervals, in-situ $\ptrue$ results, bin-count
ablations, and prompt comparisons.

\section{Confidence Estimation via MC Dropout}
We provide additional details on confidence estimation using MC Dropout. 
\subsection{Converting BALD score to $c_{\mathrm{BALD}}$}\label{app:bald}
Since BALD is an uncertainty score~\citep{gal2016dropout}, we convert token-level BALD values into confidence by applying a monotone decreasing normalization function $\phi$ and averaging over answer-token positions:
$
c_{\mathrm{BALD}}
=
\frac{1}{|\bm{y}^{\mathrm{ans}}|}
\sum_{j=1}^{|\bm{y}^{\mathrm{ans}}|}
\phi(\mathrm{BALD}_j),
$
where $\phi$ maps BALD values to $[0,1]$ with larger BALD corresponding to lower confidence.

In practice, choosing $\phi$ is non-trivial because BALD is not naturally bounded in $[0,1]$ and its scale depends on the support over which entropy is computed. We experimented with \textbf{three normalization strategies}. The first directly maps uncertainty to confidence as $\phi(\mathrm{BALD})=\mathrm{clip}(1-\mathrm{BALD},0,1)$. The second normalizes BALD over an effective top-$k$ support: for each MC Dropout sample, we identify the $k$ tokens with highest probability, take the union of these tokens across all $T$ samples, and normalize the BALD score by $\log(k_{\mathrm{eff}})$, where $k \leq k_{\mathrm{eff}} \leq Tk$. The third uses a top-$p$ support, where the effective support is determined by the smallest set of tokens whose cumulative probability exceeds a threshold $p$. This can produce a wider range of effective support sizes than the top-$k$ approach.

Across them, we found the resulting confidence scores to be sensitive to the dropout rate. This sensitivity is expected from the BALD decomposition: the predictive entropy term reflects the entropy of the averaged distribution across dropout samples and therefore increases as dropout induces more variation among predictive distributions. The expected conditional entropy term, however, remains relatively stable in our experiments, suggesting that individual dropout samples still tend to produce sharp token distributions. Therefore, changing the dropout rate primarily changes the amount of disagreement captured by the averaged predictive distribution, which directly affects the scale of BALD and motivates the need for careful normalization.

In addition to the dropout rate, the top-$k$ and top-$p$ normalization strategies introduce extra hyperparameters that must be selected. Although these approaches are more principled in that they normalize BALD with respect to an estimated effective support size, they add another source of tuning and sensitivity. For simplicity and reproducibility, we therefore use the direct clipping normalization, $\phi(\mathrm{BALD})=\mathrm{clip}(1-\mathrm{BALD},0,1)$, in our main experiments.
\begin{table}[h]
\centering
\begingroup
\scriptsize
\setlength{\tabcolsep}{4pt}
\renewcommand{\arraystretch}{0.55}

\begin{tabular}{llcccc}
\toprule
Model & Dataset & Dropout rate & ECE & MCE & Brier \\
\midrule

\multirow{14}{*}{Llama-3.1-8B}
& \multirow{3}{*}{GSM8K}
& \textbf{0.10} & \textbf{0.159} & \textbf{0.805} & \textbf{0.179} \\
& & 0.15 & 0.510 & 0.974 & 0.436 \\
& & 0.20 & 0.576 & 0.869 & 0.475 \\
\cmidrule(lr){2-6}
& \multirow{3}{*}{GSMHard}
& 0.10 & 0.308 & 0.568 & 0.340 \\
& & 0.15 & 0.143 & 0.526 & 0.249 \\
& & \textbf{0.20} & \textbf{0.141} & \textbf{0.413} & \textbf{0.230} \\
\cmidrule(lr){2-6}
& \multirow{3}{*}{SVAMP}
& \textbf{0.10} & \textbf{0.133} & \textbf{0.549} & \textbf{0.172} \\
& & 0.15 & 0.480 & 0.940 & 0.407 \\
& & 0.20 & 0.569 & 0.747 & 0.495 \\
\midrule

\multirow{14}{*}{Llama-3.2-3B}
& \multirow{3}{*}{GSM8K}
& \textbf{0.10} & \textbf{0.119} & \textbf{0.890} & \textbf{0.171} \\
& & 0.15 & 0.552 & 0.828 & 0.470 \\
& & 0.20 & 0.554 & 0.873 & 0.464 \\
\cmidrule(lr){2-6}
& \multirow{3}{*}{GSMHard}
& 0.10 & 0.279 & 0.949 & 0.250 \\
& & 0.15 & 0.149 & 0.562 & 0.245 \\
& & \textbf{0.20} & \textbf{0.106} & \textbf{0.276} & \textbf{0.205} \\
\cmidrule(lr){2-6}
& \multirow{3}{*}{SVAMP}
& \textbf{0.10} & \textbf{0.079} & \textbf{1.000} & \textbf{0.161} \\
& & 0.15 & 0.476 & 0.619 & 0.388 \\
& & 0.20 & 0.541 & 0.738 & 0.459 \\
\midrule

\multirow{14}{*}{DeepSeek-R1-8B}
& \multirow{3}{*}{GSM8K}
& \textbf{0.10} & \textbf{0.121} & \textbf{0.929} & \textbf{0.232} \\
& & 0.15 & 0.412 & 0.742 & 0.403 \\
& & 0.20 & 0.444 & 0.559 & 0.436 \\
\cmidrule(lr){2-6}
& \multirow{3}{*}{GSMHard}
& 0.10 & 0.413 & 0.809 & 0.408 \\
& & 0.15 & 0.349 & 0.726 & 0.350 \\
& & \textbf{0.20} & \textbf{0.187} & \textbf{0.723} & \textbf{0.295} \\
\cmidrule(lr){2-6}
& \multirow{3}{*}{SVAMP}
& \textbf{0.10} & \textbf{0.211} & \textbf{0.928} & \textbf{0.244} \\
& & 0.15 & 0.430 & 0.767 & 0.393 \\
& & 0.20 & 0.539 & 0.726 & 0.486 \\
\midrule

\multirow{14}{*}{Qwen3-8B-Thinking}
& \multirow{3}{*}{GSM8K}
& \textbf{0.10} & \textbf{0.044} & \textbf{0.151} & \textbf{0.041} \\
& & 0.15 & 0.119 & 0.567 & 0.064 \\
& & 0.20 & 0.335 & 0.542 & 0.167 \\
\cmidrule(lr){2-6}
& \multirow{3}{*}{GSMHard}
& 0.10 & 0.151 & 0.247 & 0.161 \\
& & \textbf{0.15} & \textbf{0.067} & \textbf{0.548} & \textbf{0.143} \\
& & 0.20 & 0.244 & 0.724 & 0.203 \\
\cmidrule(lr){2-6}
& \multirow{3}{*}{SVAMP}
& \textbf{0.10} & \textbf{0.054} & \textbf{0.250} & \textbf{0.051} \\
& & 0.15 & 0.152 & 0.547 & 0.079 \\
& & 0.20 & 0.316 & 0.825 & 0.160 \\
\midrule

\multirow{14}{*}{Qwen3-4B-Instruct}
& \multirow{3}{*}{GSM8K}
& \textbf{0.10} & \textbf{0.098} & \textbf{0.676} & \textbf{0.058} \\
& & 0.15 & 0.464 & 0.864 & 0.268 \\
& & 0.20 & 0.559 & 0.947 & 0.369 \\
\cmidrule(lr){2-6}
& \multirow{3}{*}{GSMHard}
& \textbf{0.10} & \textbf{0.155} & \textbf{0.580} & \textbf{0.226} \\
& & 0.15 & 0.198 & 0.906 & 0.243 \\
& & 0.20 & 0.294 & 0.969 & 0.313 \\
\cmidrule(lr){2-6}
& \multirow{3}{*}{SVAMP}
& \textbf{0.10} & \textbf{0.038} & \textbf{0.248} & \textbf{0.045} \\
& & 0.15 & 0.413 & 0.932 & 0.239 \\
& & 0.20 & 0.487 & 0.980 & 0.301 \\

\bottomrule
\end{tabular}
\endgroup
\caption{
\textbf{Grid search for MC-dropout rate selection.}
We report ECE, MCE, and Brier score on the selection split for each
model--dataset pair. Lower values are better.
\textbf{Bold} indicates the selected dropout rate.
}
\label{tab:dropout_grid_search}
\end{table}
\subsection{Dropout Rate Selection}\label{app:dropout}

Although fixing $\phi(\mathrm{BALD})$ reduces the hyperparameter search space, MC Dropout remains sensitive to the dropout rate. GSM8K and SVAMP provide training (calibration) and test set splits. GSMHard does not include a training split, so we randomly split the dataset into 70\% for calibration and 30\% for evaluation. From the calibration split, we randomly sample 100 examples to search over dropout rates $\{0.10,0.15,0.20\}$ for each model--dataset pair. The selected rate is then fixed and evaluated on the disjoint evaluation split in Table~\ref{tab:result_estimation}.

Table~\ref{tab:dropout_grid_search} reports performance on the 100-example tuning subset. The selected rates generally transfer well to the evaluation split, but vary across models and datasets. Notably, GSMHard favors a higher dropout rate than GSM8K and SVAMP for four of the five models, suggesting that more challenging datasets may require stronger stochastic perturbations to expose useful predictive variation. This sensitivity nevertheless introduces an additional tuning cost for MC Dropout.

\section{Implementation Details}\label{app:implementation}
In addition to the experiment setups described in Sec.~\ref{sec:exp-setup}, we provide other implementation details.

\noindent\textbf{Datasets:}
GSM8K and SVAMP provide training splits, which we use as calibration sets for fitting post-hoc calibrators, and we evaluate on their test sets. GSMHard does not include a training split, so we randomly split the dataset into 70\% for calibration and 30\% for evaluation. The resulting evaluation sets contain 1319, 400, and 300 examples for GSM8K, GSMHard, and SVAMP, respectively, while the calibration sets contain 7473, 919, and 700 examples, respectively. Although these calibration sets are relatively large, Figure~\ref{fig:calibration_analysis} shows that as few as 50 calibration examples are already effective in many settings.

\noindent\textbf{Answer extraction:} We employ a regex-based extraction method, instructing the model to provide the final numerical answer after the \texttt{\#\#\#\#} delimiter. Figure~\ref{fig:prompt_example} illustrates a representative interaction, including the system instructions, special tokens, and generated reasoning steps for Llama-3.1-8B.

Not all models were fine-tuned to follow this specific answer format, i.e., \texttt{\#\#\#\#}. In particular, DeepSeek-R1-8B was fine-tuned to produce final answers in the \texttt{\textbackslash boxed\{\}} format, and we posit that this format mismatch is the main reason for its degraded performance relative to reported results.

\noindent\textbf{One-shot Demonstration.} Our input prompt includes a one-shot demonstration (see Figure~\ref{fig:prompt_example}). We found this example to be essential for improving model accuracy and ensuring formatting consistency; specifically, it significantly increased the success rate of our regex-based answer extraction by anchoring the model to the \texttt{\#\#\#\#} delimiter.

\noindent\textbf{Calibration:} For reliability diagrams and the computation of ECE and MCE, we follow~\citet{guo2017calibration} and use 10 bins. In Table~\ref{tab:bin_ablation}, we study the effect of the number of bins on calibration for Llama-3.1-8B and Qwen3-8B-Thinking, reporting ECE, MCE, and Brier score across datasets and confidence estimators. We observe that ECE remains relatively stable as the number of bins increases, while Brier score is unchanged because it does not depend on binning. MCE is more sensitive to the binning choice. This is expected because MCE measures the largest bin-wise calibration gap: as bins become smaller, localized calibration errors are less averaged out and can produce larger maximum deviations.

\noindent\textbf{Inference:}
All inferences are performed using each model's recommended configuration, including temperature, top-$p$, top-$k$, and system prompts. In Table~\ref{tab:inference_settings}, we summarize those settings.
\begin{table*}[t]
\centering
\small
\renewcommand{\arraystretch}{1.15}
\setlength{\tabcolsep}{3.1pt}
\begin{tabular}{lccccc}
\toprule
\textbf{Setting} 
& Llama-3.1-8B 
& Llama-3.2-3B 
& DeepSeek-R1-8B 
& Qwen3-8B-Thinking 
& Qwen3-4B-Instruct \\
\midrule
Temperature & 1.0 & 0.6 & 0.6 & 0.6 & 0.7 \\
Top-$p$     & 1.0 & 0.9 & 0.95 & 0.95 & 0.8 \\
Top-$k$     & -1  & -1  & -1   & 20   & 20 \\
Min-$p$     & 0   & 0   & 0    & 0    & 0 \\
System prompt & Yes & Yes & No & Yes & Yes \\
\bottomrule
\end{tabular}
\caption{
\textbf{Recommended inference settings.}
We report the decoding and prompt settings used for each model.
}
\label{tab:inference_settings}
\end{table*}
\begin{figure}[h]
\centering
\begin{tcolorbox}[
    enhanced,
    colback=gray!4,
    colframe=gray!55,
    boxrule=0.4pt,
    arc=2pt,
    left=2pt,
    right=2pt,
    top=2pt,
    bottom=2pt,
    boxsep=1pt,
    width=\linewidth
]
\begin{lstlisting}[style=promptstyle]
<|begin_of_text|><|start_header_id|>system<|end_header_id|>

You are a helpful, respectful, and honest assistant. Always answer as helpfully as possible, while being safe. Your answers should not include any harmful, unethical, racist, sexist, toxic, dangerous, or illegal content. Please ensure that your responses are socially unbiased and positive in nature.

<|start_header_id|>user<|end_header_id|>

You are a mathematics expert. First, study this example:
Question: Sarah has 10 marbles, gives 3 away.
Answer: Sarah would have 10 - 3 = <<10-3=7>>7 marbles remaining.
#### 7

Next, solve:
Mary is baking a cake. The recipe calls for 6 cups of flour, 8 cups of sugar,
and 7 cups of salt. She already put in 5 cups of flour.
How many more cups of sugar than cups of salt does she need to add now?

At the end, write the answer as an integer after '#### '.

<|start_header_id|>assistant<|end_header_id|>

To solve this problem, let's break it down step by step:
1. The recipe calls for 8 cups of sugar and 7 cups of salt.
2. Difference = 8 - 7 = 1.

#### 1
\end{lstlisting}
\end{tcolorbox}
\caption{\textbf{Example input--output exchange} on GSM8K with Llama-3.1-8B. The one-shot example is crucial in improving the success rate of our regex-based answer extraction.}
\label{fig:prompt_example}
\end{figure}

\section{Additional Experiment Results}\label{app:additional_result}
\paragraph{Bootstrap confidence intervals.}
Tables~\ref{tab:result_estimation_full}, ~\ref{tab:result_calibration_full}, and~\ref{tab:result_isotonic} report mean estimates with 95\% percentile confidence intervals obtained from 5000 bootstrap resamples of the evaluation set. Because ECE and MCE depend on bin assignments, their bootstrap distributions may be asymmetric, and MCE intervals can be comparatively wide due to variation in the worst-calibrated bin. Nevertheless, the point estimates and confidence intervals are broadly consistent with the qualitative conclusions.

\begin{table*}[t]
\centering

\begingroup

\newcommand{\scoreci}[3]{%
    \ensuremath{#1_{\scriptscriptstyle[#2,#3]}}%
}
\newcommand{\bestscoreci}[3]{%
    \ensuremath{\mathbf{#1}_{\scriptscriptstyle[#2,#3]}}%
}

\tiny
\renewcommand{\arraystretch}{1.15}
\setlength{\tabcolsep}{3pt}

\resizebox{\textwidth}{!}{%
\begin{tabular}{llc ccc ccc ccc ccc}
\toprule
\multirow{2}{*}{Model}
& \multirow{2}{*}{Dataset}
& \multirow{2}{*}{Acc.}
& \multicolumn{3}{c}{\textsc{Seq Avg}}
& \multicolumn{3}{c}{\textsc{Seq Joint}}
& \multicolumn{3}{c}{\textsc{Ans Avg}}
& \multicolumn{3}{c}{\textsc{Ans Joint}} \\
\cmidrule(lr){4-6}
\cmidrule(lr){7-9}
\cmidrule(lr){10-12}
\cmidrule(lr){13-15}
& &
& ECE & MCE & Brier
& ECE & MCE & Brier
& ECE & MCE & Brier
& ECE & MCE & Brier \\
\midrule

\multirow{3}{*}{Llama-3.1-8B}
& GSM8K
& \scoreci{.824}{.804}{.845}
& \bestscoreci{.053}{.035}{.072}
& \scoreci{.481}{.417}{.575}
& \scoreci{.131}{.119}{.144}
& \scoreci{.117}{.101}{.137}
& \scoreci{.195}{.139}{.294}
& \scoreci{.138}{.130}{.147}
& \scoreci{.166}{.147}{.187}
& \scoreci{.529}{.357}{.818}
& \scoreci{.167}{.148}{.187}
& \scoreci{.165}{.146}{.187}
& \scoreci{.469}{.345}{.818}
& \scoreci{.167}{.147}{.187} \\

& GSMHard
& \scoreci{.350}{.302}{.398}
& \scoreci{.457}{.411}{.502}
& \scoreci{.600}{.556}{.665}
& \scoreci{.416}{.386}{.446}
& \scoreci{.320}{.277}{.363}
& \scoreci{.414}{.374}{.518}
& \scoreci{.299}{.276}{.321}
& \scoreci{.622}{.576}{.668}
& \scoreci{.743}{.709}{.856}
& \scoreci{.606}{.561}{.648}
& \scoreci{.615}{.568}{.661}
& \scoreci{.753}{.733}{.850}
& \scoreci{.596}{.551}{.639} \\

& SVAMP
& \scoreci{.827}{.780}{.870}
& \bestscoreci{.057}{.030}{.100}
& \scoreci{.309}{.148}{.492}
& \scoreci{.133}{.110}{.159}
& \scoreci{.170}{.136}{.213}
& \scoreci{.332}{.265}{.743}
& \scoreci{.160}{.143}{.178}
& \scoreci{.169}{.128}{.214}
& \scoreci{.797}{.148}{.817}
& \scoreci{.169}{.130}{.211}
& \scoreci{.171}{.130}{.215}
& \scoreci{.765}{.157}{.798}
& \scoreci{.169}{.128}{.212} \\

\midrule
\multirow{3}{*}{Llama-3.2-3B}
& GSM8K
& \scoreci{.775}{.753}{.798}
& \scoreci{.124}{.102}{.146}
& \scoreci{.587}{.289}{.793}
& \scoreci{.182}{.165}{.200}
& \scoreci{.092}{.072}{.116}
& \scoreci{.364}{.269}{.655}
& \scoreci{.173}{.157}{.189}
& \scoreci{.218}{.196}{.241}
& \scoreci{.635}{.445}{.763}
& \scoreci{.218}{.196}{.240}
& \scoreci{.218}{.197}{.240}
& \scoreci{.623}{.426}{.758}
& \scoreci{.217}{.196}{.240} \\

& GSMHard
& \scoreci{.258}{.215}{.300}
& \scoreci{.636}{.594}{.678}
& \scoreci{.773}{.751}{.791}
& \scoreci{.589}{.556}{.623}
& \scoreci{.597}{.555}{.638}
& \scoreci{.676}{.648}{.744}
& \scoreci{.538}{.506}{.569}
& \scoreci{.725}{.682}{.766}
& \scoreci{.857}{.842}{.870}
& \scoreci{.712}{.670}{.753}
& \scoreci{.722}{.677}{.766}
& \scoreci{.768}{.749}{.862}
& \scoreci{.708}{.666}{.748} \\

& SVAMP
& \scoreci{.820}{.777}{.863}
& \scoreci{.068}{.036}{.113}
& \scoreci{.783}{.773}{.792}
& \scoreci{.145}{.113}{.177}
& \bestscoreci{.035}{.016}{.081}
& \scoreci{.676}{.048}{.676}
& \scoreci{.139}{.112}{.169}
& \scoreci{.175}{.135}{.218}
& \scoreci{.840}{.734}{.871}
& \scoreci{.170}{.128}{.213}
& \scoreci{.175}{.133}{.218}
& \scoreci{.837}{.731}{.866}
& \scoreci{.170}{.129}{.212} \\

\midrule
\multirow{3}{*}{DeepSeek-R1-8B}
& GSM8K
& \scoreci{.650}{.624}{.676}
& \scoreci{.253}{.227}{.278}
& \scoreci{.384}{.341}{.427}
& \scoreci{.285}{.265}{.305}
& \scoreci{.219}{.194}{.245}
& \scoreci{.476}{.330}{.615}
& \scoreci{.267}{.249}{.286}
& \scoreci{.304}{.279}{.330}
& \scoreci{.361}{.354}{.572}
& \scoreci{.310}{.286}{.334}
& \scoreci{.302}{.277}{.329}
& \scoreci{.356}{.345}{.561}
& \scoreci{.309}{.285}{.333} \\

& GSMHard
& \scoreci{.320}{.273}{.365}
& \scoreci{.588}{.543}{.635}
& \scoreci{.646}{.578}{.718}
& \scoreci{.561}{.525}{.598}
& \scoreci{.556}{.509}{.598}
& \scoreci{.611}{.546}{.786}
& \scoreci{.523}{.487}{.556}
& \scoreci{.637}{.591}{.683}
& \scoreci{.674}{.628}{.779}
& \scoreci{.620}{.578}{.662}
& \scoreci{.631}{.583}{.676}
& \scoreci{.690}{.652}{.807}
& \scoreci{.613}{.569}{.656} \\

& SVAMP
& \scoreci{.728}{.676}{.779}
& \scoreci{.164}{.116}{.215}
& \scoreci{.784}{.155}{.784}
& \scoreci{.219}{.182}{.258}
& \bestscoreci{.127}{.078}{.180}
& \scoreci{.525}{.314}{.715}
& \scoreci{.207}{.171}{.244}
& \scoreci{.244}{.196}{.294}
& \scoreci{.617}{.603}{.789}
& \scoreci{.244}{.196}{.291}
& \scoreci{.242}{.195}{.295}
& \scoreci{.606}{.506}{.789}
& \scoreci{.242}{.196}{.290} \\

\midrule
\multirow{3}{*}{Qwen3-8B-Thinking}
& GSM8K
& \scoreci{.958}{.947}{.969}
& \bestscoreci{.006}{.001}{.017}
& \scoreci{.070}{.006}{.231}
& \scoreci{.039}{.030}{.050}
& \scoreci{.018}{.008}{.029}
& \scoreci{.027}{.012}{.080}
& \scoreci{.040}{.031}{.049}
& \scoreci{.042}{.031}{.053}
& \scoreci{.042}{.032}{.053}
& \scoreci{.042}{.031}{.053}
& \scoreci{.042}{.032}{.053}
& \scoreci{.042}{.032}{.053}
& \scoreci{.042}{.031}{.053} \\

& GSMHard
& \scoreci{.810}{.762}{.854}
& \scoreci{.144}{.101}{.189}
& \scoreci{.781}{.545}{.888}
& \scoreci{.169}{.130}{.210}
& \scoreci{.128}{.085}{.172}
& \scoreci{.793}{.287}{.793}
& \scoreci{.163}{.126}{.203}
& \scoreci{.190}{.149}{.238}
& \scoreci{.190}{.149}{.235}
& \scoreci{.190}{.146}{.235}
& \scoreci{.190}{.149}{.234}
& \scoreci{.190}{.146}{.238}
& \scoreci{.190}{.146}{.235} \\

& SVAMP
& \scoreci{.972}{.951}{.990}
& \bestscoreci{.017}{.003}{.036}
& \scoreci{.017}{.011}{.281}
& \scoreci{.026}{.011}{.044}
& \scoreci{.033}{.018}{.052}
& \scoreci{.034}{.023}{.193}
& \scoreci{.027}{.012}{.044}
& \scoreci{.028}{.010}{.049}
& \scoreci{.028}{.010}{.049}
& \scoreci{.028}{.010}{.049}
& \scoreci{.028}{.010}{.049}
& \scoreci{.028}{.010}{.049}
& \scoreci{.028}{.010}{.049} \\

\midrule
\multirow{3}{*}{Qwen3-4B-Instruct}
& GSM8K
& \scoreci{.938}{.925}{.951}
& \scoreci{.035}{.023}{.048}
& \scoreci{.478}{.151}{.777}
& \scoreci{.057}{.045}{.069}
& \bestscoreci{.028}{.016}{.041}
& \scoreci{.790}{.176}{.792}
& \scoreci{.056}{.045}{.067}
& \scoreci{.062}{.049}{.075}
& \scoreci{.062}{.049}{.075}
& \scoreci{.062}{.049}{.075}
& \scoreci{.062}{.049}{.075}
& \scoreci{.062}{.049}{.075}
& \scoreci{.062}{.049}{.075} \\

& GSMHard
& \scoreci{.701}{.656}{.744}
& \scoreci{.266}{.220}{.310}
& \scoreci{.724}{.575}{.845}
& \scoreci{.267}{.225}{.309}
& \scoreci{.256}{.211}{.301}
& \scoreci{.787}{.548}{.793}
& \scoreci{.258}{.219}{.298}
& \scoreci{.298}{.253}{.346}
& \scoreci{.299}{.254}{.345}
& \scoreci{.298}{.251}{.346}
& \scoreci{.298}{.253}{.344}
& \scoreci{.299}{.252}{.347}
& \scoreci{.298}{.251}{.346} \\

& SVAMP
& \scoreci{.943}{.913}{.967}
& \scoreci{.032}{.010}{.059}
& \scoreci{.366}{.036}{.733}
& \scoreci{.051}{.030}{.074}
& \bestscoreci{.025}{.008}{.052}
& \scoreci{.267}{.093}{.753}
& \scoreci{.049}{.029}{.074}
& \scoreci{.057}{.033}{.084}
& \scoreci{.057}{.030}{.084}
& \scoreci{.057}{.033}{.084}
& \scoreci{.057}{.033}{.084}
& \scoreci{.057}{.033}{.084}
& \scoreci{.057}{.033}{.084} \\

\bottomrule
\end{tabular}%
}

\vspace{6pt}

\textbf{(a) Single-pass estimators}

\vspace{2pt}

\resizebox{0.67\textwidth}{!}{%
\begin{tabular}{ll ccc ccc}
\toprule
\multirow{2}{*}{Model}
& \multirow{2}{*}{Dataset}
& \multicolumn{3}{c}{$\ptrue$}
& \multicolumn{3}{c}{$c_{\mathrm{BALD}}$} \\
\cmidrule(lr){3-5}
\cmidrule(lr){6-8}
& & ECE & MCE & Brier
& ECE & MCE & Brier \\
\midrule

\multirow{3}{*}{Llama-3.1-8B}
& GSM8K
& \scoreci{.097}{.083}{.120}
& \scoreci{.625}{.228}{.893}
& \scoreci{.143}{.127}{.159}
& \scoreci{.149}{.129}{.170}
& \scoreci{.667}{.521}{1.000}
& \scoreci{.182}{.171}{.194} \\

& GSMHard
& \scoreci{.333}{.294}{.381}
& \scoreci{.627}{.556}{.720}
& \scoreci{.319}{.289}{.350}
& \bestscoreci{.152}{.118}{.203}
& \scoreci{.424}{.256}{.558}
& \scoreci{.252}{.222}{.282} \\

& SVAMP
& \scoreci{.099}{.071}{.147}
& \scoreci{.450}{.440}{.972}
& \scoreci{.139}{.108}{.173}
& \scoreci{.095}{.074}{.147}
& \scoreci{.633}{.621}{1.000}
& \scoreci{.158}{.134}{.184} \\

\midrule
\multirow{3}{*}{Llama-3.2-3B}
& GSM8K
& \bestscoreci{.055}{.039}{.077}
& \scoreci{.304}{.138}{.634}
& \scoreci{.161}{.150}{.173}
& \scoreci{.108}{.089}{.132}
& \scoreci{.389}{.302}{.651}
& \scoreci{.186}{.173}{.200} \\

& GSMHard
& \scoreci{.251}{.221}{.295}
& \scoreci{.406}{.356}{.542}
& \scoreci{.229}{.210}{.248}
& \bestscoreci{.078}{.049}{.125}
& \scoreci{.202}{.126}{.512}
& \scoreci{.196}{.171}{.221} \\

& SVAMP
& \scoreci{.104}{.078}{.149}
& \scoreci{.423}{.326}{.828}
& \scoreci{.148}{.125}{.173}
& \scoreci{.097}{.072}{.149}
& \scoreci{.610}{.427}{.896}
& \scoreci{.167}{.142}{.194} \\

\midrule
\multirow{3}{*}{DeepSeek-R1-8B}
& GSM8K
& \scoreci{.315}{.289}{.341}
& \scoreci{.484}{.326}{.972}
& \scoreci{.326}{.303}{.350}
& \bestscoreci{.168}{.142}{.195}
& \scoreci{.732}{.510}{.937}
& \scoreci{.265}{.252}{.278} \\

& GSMHard
& \scoreci{.634}{.588}{.680}
& \scoreci{.673}{.664}{.817}
& \scoreci{.618}{.576}{.661}
& \bestscoreci{.102}{.075}{.154}
& \scoreci{.606}{.443}{.610}
& \scoreci{.239}{.214}{.265} \\

& SVAMP
& \scoreci{.241}{.194}{.293}
& \scoreci{.536}{.360}{.680}
& \scoreci{.252}{.207}{.299}
& \scoreci{.174}{.133}{.231}
& \scoreci{.979}{.361}{.979}
& \scoreci{.240}{.214}{.270} \\

\midrule
\multirow{3}{*}{Qwen3-8B-Thinking}
& GSM8K
& \scoreci{.035}{.025}{.047}
& \scoreci{.829}{.246}{.829}
& \scoreci{.042}{.032}{.053}
& \scoreci{.043}{.032}{.054}
& \scoreci{.545}{.193}{.692}
& \scoreci{.042}{.032}{.053} \\

& GSMHard
& \scoreci{.183}{.141}{.229}
& \scoreci{.373}{.179}{.897}
& \scoreci{.183}{.141}{.227}
& \bestscoreci{.028}{.022}{.082}
& \scoreci{.365}{.107}{.665}
& \scoreci{.148}{.122}{.175} \\

& SVAMP
& \scoreci{.020}{.005}{.041}
& \scoreci{.348}{.123}{.773}
& \scoreci{.027}{.011}{.047}
& \scoreci{.026}{.010}{.048}
& \scoreci{.133}{.110}{.260}
& \scoreci{.026}{.011}{.048} \\

\midrule
\multirow{3}{*}{Qwen3-4B-Instruct}
& GSM8K
& \scoreci{.062}{.050}{.076}
& \scoreci{.652}{.626}{1.000}
& \scoreci{.063}{.050}{.077}
& \scoreci{.061}{.048}{.075}
& \scoreci{.650}{.624}{.697}
& \scoreci{.069}{.058}{.080} \\

& GSMHard
& \scoreci{.285}{.242}{.332}
& \scoreci{.985}{.453}{.985}
& \scoreci{.286}{.242}{.331}
& \bestscoreci{.188}{.150}{.241}
& \scoreci{.451}{.315}{.464}
& \scoreci{.248}{.212}{.284} \\

& SVAMP
& \scoreci{.058}{.033}{.087}
& \scoreci{.999}{.109}{.999}
& \scoreci{.060}{.034}{.088}
& \scoreci{.061}{.039}{.091}
& \scoreci{.313}{.246}{.477}
& \scoreci{.061}{.040}{.086} \\

\bottomrule

\end{tabular}%
}

\vspace{2pt}

\textbf{(b) Multi-pass estimators}

\vspace{2pt}

\endgroup
\caption{
\textbf{Full confidence-estimation results using token probabilities.}
We report point estimates with 95\% percentile confidence intervals from
5,000 bootstrap resamples shown as subscripts. Lower values are better for
ECE, MCE, and Brier score. Bold indicates the lowest ECE point estimate
across all estimators for each model--dataset pair.
}
\label{tab:result_estimation_full}

\end{table*}
\begin{table*}[t]
\centering
\begingroup

\newcommand{\calci}[4]{%
    \ensuremath{#1_{\scriptscriptstyle[#2,#3]}\,(#4)}%
}
\newcommand{\bestcalci}[4]{%
    \ensuremath{\mathbf{#1}_{\scriptscriptstyle[#2,#3]}\,
    \mathbf{(#4)}}%
}

\renewcommand{\arraystretch}{0.9}
\setlength{\tabcolsep}{3.2pt}

\scriptsize
\resizebox{\textwidth}{!}{%
\begin{tabular}{ll | ccc | ccc cc}
\toprule
\multirow{2}{*}{Model}
& \multirow{2}{*}{Dataset}
& \multicolumn{3}{c|}{Isotonic Regression}
& \multicolumn{5}{c}{Platt Scaling} \\
\cmidrule(lr){3-5}
\cmidrule(lr){6-10}
&
& \textsc{Ans Avg}
& \textsc{Ans Joint}
& $\ptrue$
& $t$
& \textsc{Ans Avg}
& \textsc{Ans Joint}
& $(a,b)$
& $\ptrue$ \\
\midrule

\multirow{3}{*}{Llama-3.1-8B}
& GSM8K
& \bestcalci{0.056}{0.038}{0.076}{+66\%}
& \bestcalci{0.054}{0.036}{0.075}{+67\%}
& \bestcalci{0.058}{0.041}{0.079}{+40\%}
& 1.82
& \calci{0.081}{0.063}{0.101}{+52\%}
& \calci{0.080}{0.064}{0.101}{+51\%}
& $(0.53,0.48)$
& \calci{0.059}{0.041}{0.079}{+39\%} \\

& GSMHard
& \bestcalci{0.064}{0.036}{0.110}{+90\%}
& \calci{0.065}{0.038}{0.110}{+89\%}
& \bestcalci{0.057}{0.037}{0.105}{+83\%}
& 2.37
& \calci{0.091}{0.062}{0.141}{+85\%}
& \bestcalci{0.045}{0.036}{0.099}{+93\%}
& $(0.57,-1.74)$
& \calci{0.087}{0.063}{0.131}{+74\%} \\

& SVAMP
& \bestcalci{0.020}{0.019}{0.070}{+88\%}
& \bestcalci{0.022}{0.021}{0.070}{+87\%}
& \calci{0.045}{0.029}{0.093}{+54\%}
& 1.91
& \calci{0.084}{0.055}{0.125}{+50\%}
& \calci{0.083}{0.054}{0.123}{+52\%}
& $(0.54,0.43)$
& \bestcalci{0.039}{0.027}{0.089}{+60\%} \\

\midrule

\multirow{3}{*}{Llama-3.2-3B}
& GSM8K
& \bestcalci{0.074}{0.054}{0.096}{+66\%}
& \bestcalci{0.075}{0.056}{0.097}{+65\%}
& \calci{0.076}{0.056}{0.098}{-39\%}
& 1.74
& \calci{0.104}{0.084}{0.127}{+52\%}
& \calci{0.103}{0.082}{0.124}{+53\%}
& $(0.65,1.07)$
& \bestcalci{0.044}{0.034}{0.096}{+20\%} \\

& GSMHard
& \bestcalci{0.027}{0.017}{0.073}{+96\%}
& \bestcalci{0.028}{0.017}{0.075}{+96\%}
& \bestcalci{0.031}{0.028}{0.077}{+88\%}
& 2.27
& \calci{0.105}{0.079}{0.150}{+85\%}
& \calci{0.057}{0.037}{0.102}{+92\%}
& $(0.79,-1.36)$
& \calci{0.047}{0.036}{0.091}{+81\%} \\

& SVAMP
& \bestcalci{0.036}{0.030}{0.081}{+79\%}
& \bestcalci{0.036}{0.029}{0.081}{+80\%}
& \calci{0.055}{0.034}{0.099}{+48\%}
& 1.79
& \calci{0.044}{0.031}{0.090}{+75\%}
& \calci{0.041}{0.029}{0.087}{+76\%}
& $(0.61,0.80)$
& \bestcalci{0.029}{0.025}{0.083}{+72\%} \\

\midrule

\multirow{3}{*}{DeepSeek-R1-8B}
& GSM8K
& \bestcalci{0.012}{0.008}{0.042}{+96\%}
& \bestcalci{0.013}{0.007}{0.042}{+96\%}
& \calci{0.017}{0.006}{0.044}{+95\%}
& 1.68
& \calci{0.071}{0.053}{0.099}{+77\%}
& \calci{0.070}{0.053}{0.099}{+77\%}
& $(0.054,0.46)$
& \bestcalci{0.016}{0.003}{0.043}{+95\%} \\

& GSMHard
& \bestcalci{0.068}{0.037}{0.115}{+89\%}
& \bestcalci{0.071}{0.040}{0.120}{+89\%}
& \calci{0.066}{0.031}{0.117}{+90\%}
& 2.15
& \calci{0.100}{0.080}{0.154}{+84\%}
& \calci{0.082}{0.057}{0.134}{+87\%}
& $(0.065,-1.29)$
& \bestcalci{0.065}{0.020}{0.110}{+90\%} \\

& SVAMP
& \bestcalci{0.016}{0.015}{0.076}{+93\%}
& \bestcalci{0.024}{0.019}{0.082}{+90\%}
& \calci{0.034}{0.014}{0.087}{+86\%}
& 1.67
& \calci{0.061}{0.044}{0.122}{+75\%}
& \calci{0.066}{0.046}{0.126}{+73\%}
& $(0.022,0.83)$
& \bestcalci{0.013}{0.002}{0.064}{+95\%} \\

\midrule

\multirow{3}{*}{Qwen3-8B-Thinking}
& GSM8K
& \bestcalci{0.015}{0.005}{0.027}{+63\%}
& \bestcalci{0.015}{0.005}{0.027}{+63\%}
& \bestcalci{0.015}{0.004}{0.026}{+58\%}
& 3.19
& \calci{0.030}{0.020}{0.042}{+29\%}
& \calci{0.030}{0.020}{0.042}{+28\%}
& $(0.39,1.54)$
& \calci{0.016}{0.006}{0.028}{+54\%} \\

& GSMHard
& \bestcalci{0.034}{0.009}{0.077}{+82\%}
& \bestcalci{0.033}{0.009}{0.076}{+83\%}
& \bestcalci{0.049}{0.028}{0.094}{+73\%}
& 3.62
& \calci{0.114}{0.078}{0.161}{+40\%}
& \calci{0.114}{0.079}{0.162}{+40\%}
& $(0.52,-1.38)$
& \calci{0.078}{0.055}{0.124}{+57\%} \\

& SVAMP
& \bestcalci{0.003}{0.001}{0.022}{+91\%}
& \bestcalci{0.003}{0.001}{0.022}{+91\%}
& \calci{0.007}{0.003}{0.026}{+66\%}
& 3.22
& \calci{0.022}{0.005}{0.042}{+20\%}
& \calci{0.022}{0.005}{0.043}{+21\%}
& $(0.46,1.22)$
& \bestcalci{0.004}{0.001}{0.024}{+79\%} \\

\midrule

\multirow{3}{*}{Qwen3-4B-Instruct}
& GSM8K
& \bestcalci{0.019}{0.006}{0.033}{+69\%}
& \bestcalci{0.019}{0.006}{0.033}{+69\%}
& \calci{0.020}{0.008}{0.033}{+68\%}
& 3.90
& \calci{0.044}{0.031}{0.057}{+29\%}
& \calci{0.044}{0.031}{0.057}{+29\%}
& $(0.18,1.70)$
& \bestcalci{0.019}{0.007}{0.032}{+70\%} \\

& GSMHard
& \bestcalci{0.077}{0.032}{0.123}{+74\%}
& \bestcalci{0.077}{0.032}{0.124}{+74\%}
& \bestcalci{0.091}{0.060}{0.140}{+68\%}
& 4.48
& \calci{0.131}{0.097}{0.181}{+56\%}
& \calci{0.144}{0.109}{0.193}{+52\%}
& $(0.11,-0.18)$
& \calci{0.092}{0.067}{0.146}{+68\%} \\

& SVAMP
& \bestcalci{0.035}{0.012}{0.062}{+38\%}
& \bestcalci{0.035}{0.012}{0.062}{+38\%}
& \calci{0.032}{0.009}{0.059}{+44\%}
& 3.88
& \calci{0.043}{0.020}{0.070}{+24\%}
& \calci{0.044}{0.020}{0.071}{+22\%}
& $(-0.30,6.02)$
& \bestcalci{0.032}{0.008}{0.060}{+45\%} \\

\bottomrule
\end{tabular}%
}

\caption{
\textbf{Post-hoc calibration results.}
We report calibrated ECE with 95\% percentile bootstrap confidence intervals
shown as subscripts for \textsc{Ans Avg}, \textsc{Ans Joint}, and $\ptrue$
after isotonic regression and Platt scaling. Relative improvements over the
corresponding uncalibrated estimators are shown in parentheses. For Platt
scaling, we also report the learned temperature $t$ for answer-based
estimators and the learned parameters $(a,b)$ for $\ptrue$. For each
estimator and dataset, \textbf{bold} indicates the lower ECE between the two
calibration methods.
}
\label{tab:result_calibration_full}

\endgroup
\end{table*}
\begin{table*}[t]
\centering

\begingroup
\newcommand{\isoCI}[4]{%
    \ensuremath{#1_{\scriptscriptstyle[#2,#3]}}%
    \hspace{0.15em}\ensuremath{(#4)}%
}
\newcommand{\isoBestCI}[4]{%
    \ensuremath{\mathbf{#1}_{\scriptscriptstyle[#2,#3]}}%
    \hspace{0.15em}\textbf{\ensuremath{(#4)}}%
}

\renewcommand{\arraystretch}{1.3}
\setlength{\tabcolsep}{4pt}
\scriptsize

% ================================================================
% Panel (a): Single-pass estimators
% ================================================================
\resizebox{\textwidth}{!}{%
\begin{tabular}{llcccc}
\toprule
Model
& Dataset
& \textsc{Seq Avg}
& \textsc{Seq Joint}
& \textsc{Ans Avg}
& \textsc{Ans Joint} \\
\midrule

\multirow{3}{*}{Llama-3.1-8B}
& GSM8K
& \isoCI{0.054}{0.036}{0.073}{-1\%}
& \isoCI{0.056}{0.039}{0.076}{+52\%}
& \isoCI{0.056}{0.038}{0.076}{+66\%}
& \isoBestCI{0.054}{0.036}{0.075}{+67\%} \\

& GSMHard
& \isoCI{0.055}{0.029}{0.104}{+88\%}
& \isoCI{0.060}{0.035}{0.107}{+81\%}
& \isoBestCI{0.064}{0.036}{0.110}{+90\%}
& \isoCI{0.065}{0.038}{0.110}{+89\%} \\

& SVAMP
& \isoCI{0.033}{0.023}{0.081}{+42\%}
& \isoCI{0.030}{0.017}{0.080}{+82\%}
& \isoBestCI{0.020}{0.019}{0.070}{+88\%}
& \isoCI{0.022}{0.021}{0.070}{+87\%} \\

\midrule
\multirow{3}{*}{Llama-3.2-3B}
& GSM8K
& \isoCI{0.074}{0.054}{0.096}{+40\%}
& \isoCI{0.074}{0.053}{0.096}{+20\%}
& \isoBestCI{0.074}{0.054}{0.096}{+66\%}
& \isoCI{0.075}{0.056}{0.097}{+65\%} \\

& GSMHard
& \isoCI{0.032}{0.021}{0.080}{+95\%}
& \isoCI{0.042}{0.030}{0.091}{+93\%}
& \isoBestCI{0.027}{0.017}{0.073}{+96\%}
& \isoBestCI{0.028}{0.017}{0.075}{+96\%} \\

& SVAMP
& \isoCI{0.029}{0.019}{0.075}{+57\%}
& \isoCI{0.040}{0.022}{0.083}{-13\%}
& \isoCI{0.036}{0.030}{0.081}{+79\%}
& \isoBestCI{0.036}{0.029}{0.081}{+80\%} \\

\midrule
\multirow{3}{*}{DeepSeek-R1-8B}
& GSM8K
& \isoCI{0.032}{0.021}{0.060}{+88\%}
& \isoCI{0.031}{0.022}{0.062}{+86\%}
& \isoBestCI{0.012}{0.008}{0.042}{+96\%}
& \isoBestCI{0.013}{0.007}{0.042}{+96\%} \\

& GSMHard
& \isoCI{0.060}{0.032}{0.110}{+90\%}
& \isoBestCI{0.053}{0.027}{0.101}{+91\%}
& \isoCI{0.068}{0.037}{0.115}{+89\%}
& \isoCI{0.071}{0.040}{0.120}{+89\%} \\

& SVAMP
& \isoCI{0.049}{0.030}{0.101}{+70\%}
& \isoCI{0.068}{0.037}{0.117}{+47\%}
& \isoBestCI{0.016}{0.015}{0.076}{+93\%}
& \isoCI{0.024}{0.019}{0.082}{+90\%} \\

\midrule
\multirow{3}{*}{Qwen3-8B-Thinking}
& GSM8K
& \isoCI{0.007}{0.006}{0.029}{-15\%}
& \isoCI{0.017}{0.007}{0.029}{+8\%}
& \isoCI{0.015}{0.005}{0.027}{+63\%}
& \isoCI{0.015}{0.005}{0.027}{+63\%} \\

& GSMHard
& \isoBestCI{0.019}{0.014}{0.067}{+87\%}
& \isoCI{0.036}{0.023}{0.082}{+71\%}
& \isoCI{0.034}{0.009}{0.077}{+82\%}
& \isoCI{0.033}{0.009}{0.076}{+83\%} \\

& SVAMP
& \isoCI{0.005}{0.004}{0.027}{+70\%}
& \isoCI{0.008}{0.005}{0.031}{+75\%}
& \isoBestCI{0.003}{0.001}{0.022}{+91\%}
& \isoBestCI{0.003}{0.001}{0.022}{+91\%} \\

\midrule
\multirow{3}{*}{Qwen3-4B-Instruct}
& GSM8K
& \isoCI{0.014}{0.006}{0.027}{+60\%}
& \isoCI{0.012}{0.005}{0.025}{+59\%}
& \isoBestCI{0.019}{0.006}{0.033}{+69\%}
& \isoBestCI{0.019}{0.006}{0.033}{+69\%} \\

& GSMHard
& \isoBestCI{0.050}{0.033}{0.095}{+81\%}
& \isoCI{0.052}{0.041}{0.099}{+80\%}
& \isoCI{0.077}{0.032}{0.123}{+74\%}
& \isoCI{0.077}{0.032}{0.124}{+74\%} \\

& SVAMP
& \isoCI{0.036}{0.015}{0.061}{-12\%}
& \isoCI{0.035}{0.014}{0.060}{-38\%}
& \isoCI{0.035}{0.012}{0.062}{+38\%}
& \isoCI{0.035}{0.012}{0.062}{+38\%} \\

\bottomrule
\end{tabular}%
}

\vspace{3pt}

\textbf{(a) Single-pass estimators}

\vspace{8pt}

% ================================================================
% Panel (b): Multi-pass estimators
% ================================================================
\makebox[\textwidth][c]{%
\resizebox{0.66\textwidth}{!}{%
\begin{tabular}{llcc}
\toprule
Model
& Dataset
& $\ptrue$
& $c_{\mathrm{BALD}}$ \\
\midrule

\multirow{3}{*}{Llama-3.1-8B}
& GSM8K
& \isoCI{0.058}{0.041}{0.079}{+40\%}
& \isoCI{0.062}{0.043}{0.084}{+58\%} \\

& GSMHard
& \isoCI{0.057}{0.037}{0.105}{+83\%}
& \isoCI{0.122}{0.080}{0.170}{+20\%} \\

& SVAMP
& \isoCI{0.045}{0.029}{0.093}{+54\%}
& \isoCI{0.040}{0.025}{0.086}{+58\%} \\

\midrule
\multirow{3}{*}{Llama-3.2-3B}
& GSM8K
& \isoCI{0.076}{0.056}{0.098}{-39\%}
& \isoCI{0.048}{0.026}{0.072}{+56\%} \\

& GSMHard
& \isoCI{0.031}{0.028}{0.077}{+88\%}
& \isoCI{0.067}{0.021}{0.105}{+14\%} \\

& SVAMP
& \isoCI{0.055}{0.034}{0.099}{+48\%}
& \isoCI{0.060}{0.027}{0.103}{+38\%} \\

\midrule
\multirow{3}{*}{DeepSeek-R1-8B}
& GSM8K
& \isoCI{0.017}{0.006}{0.044}{+95\%}
& \isoCI{0.095}{0.075}{0.123}{+43\%} \\

& GSMHard
& \isoCI{0.066}{0.031}{0.117}{+90\%}
& \isoCI{0.088}{0.056}{0.133}{+14\%} \\

& SVAMP
& \isoCI{0.034}{0.014}{0.087}{+86\%}
& \isoCI{0.053}{0.016}{0.110}{+70\%} \\

\midrule
\multirow{3}{*}{Qwen3-8B-Thinking}
& GSM8K
& \isoCI{0.015}{0.004}{0.026}{+58\%}
& \isoBestCI{0.004}{0.000}{0.015}{+92\%} \\

& GSMHard
& \isoCI{0.049}{0.028}{0.094}{+73\%}
& \isoCI{0.032}{0.015}{0.081}{-14\%} \\

& SVAMP
& \isoCI{0.007}{0.003}{0.026}{+66\%}
& \isoCI{0.031}{0.024}{0.055}{-20\%} \\

\midrule
\multirow{3}{*}{Qwen3-4B-Instruct}
& GSM8K
& \isoCI{0.020}{0.008}{0.033}{+68\%}
& \isoCI{0.032}{0.020}{0.045}{+47\%} \\

& GSMHard
& \isoCI{0.091}{0.060}{0.140}{+68\%}
& \isoCI{0.088}{0.051}{0.137}{+53\%} \\

& SVAMP
& \isoCI{0.032}{0.009}{0.059}{+44\%}
& \isoBestCI{0.029}{0.010}{0.058}{+53\%} \\

\bottomrule
\end{tabular}%
}}

\vspace{3pt}

\textbf{(b) Multi-pass estimators}

\caption{
\textbf{Calibration performance after applying isotonic regression to
confidence estimators.}
We report ECE with 95\% percentile bootstrap confidence intervals shown as
subscripts and relative improvements over the corresponding uncalibrated
estimators shown in parentheses. Bold indicates the largest relative
improvement across all estimators for each model--dataset pair; ties at the
reported precision are all bolded.
}
\label{tab:result_isotonic}

\endgroup
\end{table*}

\paragraph{Cross-domain calibration transfer.}
Table~\ref{tab:calibration_transfer_full} expands the transfer results in Table~\ref{tab:calibration_transfer_summary} by reporting ECE, MCE, and Brier for each source--target pair. 
We notice that cross-domain calibration is strongly asymmetric: mappings fitted on GSM8K improve calibration on GSMHard for both models and calibration methods, whereas transfer from GSMHard to GSM8K generally degrades calibration. 

Similarly, cross-model transfer is also target-dependent. On GSMHard, transferring between Llama and Qwen improves ECE in both directions, whereas on GSM8K, transfer from Llama to Qwen produces little change and the reverse direction substantially worsens calibration. 

\paragraph{In-situ self-verification.}
We compare in-situ $\ptrue$ with standard
$\ptrue$ and \textsc{Ans Avg} in Table~\ref{tab:insitu_ptrue}. Compared with standard $\ptrue$, it matches or improves ECE in 10 of the 15 settings.

\begin{table*}[t]
\centering
\begingroup

\renewcommand{\arraystretch}{1.08}
\setlength{\tabcolsep}{2.7pt}
\tiny

\resizebox{\textwidth}{!}{%
\begin{tabular}{ll ccc ccc ccc ccc}
\toprule
\multirow{2}{*}{Calibration}
& \multirow{2}{*}{Source $\backslash$ Target}
& \multicolumn{3}{c}{Llama / GSM8K}
& \multicolumn{3}{c}{Llama / GSMHard}
& \multicolumn{3}{c}{Qwen / GSM8K}
& \multicolumn{3}{c}{Qwen / GSMHard} \\
\cmidrule(lr){3-5}
\cmidrule(lr){6-8}
\cmidrule(lr){9-11}
\cmidrule(lr){12-14}
& & ECE & MCE & Brier
& ECE & MCE & Brier
& ECE & MCE & Brier
& ECE & MCE & Brier \\
\midrule

Uncalibrated
& --
& 0.166 & 0.529 & 0.167
& 0.622 & 0.743 & 0.606
& 0.062 & 0.062 & 0.062
& 0.298 & 0.299 & 0.298 \\

\midrule

\multirow{4}{*}{Isotonic}
& Llama / GSM8K
& \textbf{0.056} & \textbf{0.500} & \textbf{0.134}
& 0.413 & 0.490 & 0.362
& 0.062 & 0.062 & 0.062
& 0.293 & 0.295 & 0.295 \\

& Llama / GSMHard
& 0.355 & 0.645 & 0.263
& \textbf{0.064} & \textbf{0.163} & \textbf{0.187}
& 0.341 & 0.341 & 0.174
& 0.113 & 0.653 & 0.224 \\

& Qwen / GSM8K
& 0.624 & 0.624 & 0.534
& 0.150 & 0.150 & 0.250
& \textbf{0.019} & \textbf{0.800} & \textbf{0.059}
& 0.264 & 0.789 & 0.279 \\

& Qwen / GSMHard
& 0.345 & 0.361 & 0.255
& 0.095 & 0.168 & 0.204
& 0.304 & 0.500 & 0.150
& \textbf{0.077} & \textbf{0.288} & \textbf{0.217} \\

\midrule

\multirow{4}{*}{Platt}
& Llama / GSM8K
& \textbf{0.080} & \textbf{0.213} & \textbf{0.134}
& 0.539 & 0.663 & 0.483
& 0.062 & 0.586 & 0.062
& 0.298 & 0.298 & 0.298 \\

& Llama / GSMHard
& 0.379 & 0.445 & 0.264
& \textbf{0.091} & \textbf{0.175} & \textbf{0.219}
& 0.062 & 0.567 & 0.062
& 0.295 & 0.874 & 0.295 \\

& Qwen / GSM8K
& 0.824 & 0.824 & 0.815
& 0.295 & 0.295 & 0.299
& \textbf{0.044} & \textbf{0.716} & \textbf{0.059}
& 0.244 & 0.759 & 0.260 \\

& Qwen / GSMHard
& 0.830 & 0.830 & 0.827
& 0.304 & 0.304 & 0.306
& 0.051 & 0.824 & 0.058
& \textbf{0.131} & \textbf{0.214} & \textbf{0.210} \\

\bottomrule
\end{tabular}%
}

\caption{
\textbf{Cross-domain calibration transfer for \textsc{Ans Avg}.}
Each calibrator is fitted on the source model--dataset pair shown in the rows
and applied without refitting to the target pair shown in the columns.
We report ECE, MCE, and Brier score; lower values are better.
Bold indicates in-domain calibration, where the source and target pairs match.
Llama denotes Llama-3.1-8B, and Qwen denotes Qwen3-4B-Instruct.
}
\label{tab:calibration_transfer_full}

\endgroup
\end{table*}

\begin{table*}[t]
\centering
\scriptsize
\renewcommand{\arraystretch}{1.15}
\setlength{\tabcolsep}{5pt}

\begin{tabular}{llcccc}
\toprule
\textbf{Model} & \textbf{Dataset} & \textbf{Ans Avg} & \textbf{In-situ $\ptrue$} & \textbf{$\ptrue$} & \textbf{Reduction (\%)} \\
\midrule
\multirow{3}{*}{Llama-3.1-8B}
& GSM8K   & 0.166 & 0.030 & 0.097 & 88.75 \\
& GSMHard & 0.622 & 0.465 & 0.333 & 89.50 \\
& SVAMP   & 0.169 & 0.061 & 0.099 & 86.13 \\
\midrule
\multirow{3}{*}{Llama-3.2-3B}
& GSM8K   & 0.218 & 0.092 & 0.055 & 86.58 \\
& GSMHard & 0.725 & 0.481 & 0.251 & 89.01 \\
& SVAMP   & 0.175 & 0.154 & 0.104 & 85.50 \\
\midrule
\multirow{3}{*}{DeepSeek-R1-8B}
& GSM8K   & 0.304 & 0.270 & 0.315 & 91.43 \\
& GSMHard & 0.637 & 0.608 & 0.634 & 91.92 \\
& SVAMP   & 0.244 & 0.192 & 0.241 & 89.36 \\
\midrule
\multirow{3}{*}{Qwen3-8B-Thinking}
& GSM8K   & 0.042 & 0.027 & 0.035 & 91.27 \\
& GSMHard & 0.190 & 0.168 & 0.183 & 93.08 \\
& SVAMP   & 0.028 & 0.010 & 0.020 & 89.48 \\
\midrule
\multirow{3}{*}{Qwen3-4B-Instruct}
& GSM8K   & 0.062 & 0.062 & 0.062 & 83.48 \\
& GSMHard & 0.298 & 0.287 & 0.285 & 86.87 \\
& SVAMP   & 0.057 & 0.056 & 0.058 & 78.77 \\
\bottomrule
\end{tabular}
\caption{
\textbf{In-situ $\ptrue$ calibration and token-processing reduction.}
We compare \textsc{Ans Avg}, in-situ $\ptrue$, and standard $\ptrue$ in terms of ECE. The last column reports the percentage reduction in token-processing overhead for in-situ $\ptrue$ relative to standard $\ptrue$.
}
\label{tab:insitu_ptrue}
\end{table*}

\begin{table*}[t]
\centering
\scriptsize
\renewcommand{\arraystretch}{0.8}
\setlength{\tabcolsep}{3.5pt}

\resizebox{\textwidth}{!}{%
\begin{tabular}{lllccc ccc ccc ccc}
\toprule
\multirow{2}{*}{Model}
& \multirow{2}{*}{Dataset}
& \multirow{2}{*}{Bins}
& \multicolumn{3}{c}{\textsc{Seq Avg}}
& \multicolumn{3}{c}{\textsc{Seq Joint}}
& \multicolumn{3}{c}{\textsc{Ans Avg}}
& \multicolumn{3}{c}{\textsc{Ans Joint}} \\
\cmidrule(lr){4-6}
\cmidrule(lr){7-9}
\cmidrule(lr){10-12}
\cmidrule(lr){13-15}
& & & ECE & MCE & Brier & ECE & MCE & Brier & ECE & MCE & Brier & ECE & MCE & Brier \\
\midrule

\multirow{9}{*}{Llama-3.1-8B}
& \multirow{3}{*}{GSM8K}
& 10 & 0.053 & 0.481 & 0.131 & 0.117 & 0.195 & 0.138 & 0.166 & 0.529 & 0.167 & 0.165 & 0.469 & 0.167 \\
& & 20 & 0.053 & 0.516 & 0.131 & 0.123 & 0.275 & 0.138 & 0.167 & 0.641 & 0.167 & 0.165 & 0.469 & 0.167 \\
& & 30 & 0.054 & 0.555 & 0.131 & 0.122 & 0.312 & 0.138 & 0.167 & 0.641 & 0.167 & 0.166 & 0.518 & 0.167 \\
\cmidrule(lr){2-15}
& \multirow{3}{*}{GSMHard}
& 10 & 0.457 & 0.600 & 0.416 & 0.320 & 0.414 & 0.299 & 0.622 & 0.743 & 0.606 & 0.615 & 0.753 & 0.596 \\
& & 20 & 0.457 & 0.683 & 0.416 & 0.320 & 0.526 & 0.299 & 0.622 & 0.828 & 0.606 & 0.615 & 0.832 & 0.596 \\
& & 30 & 0.457 & 0.720 & 0.416 & 0.320 & 0.549 & 0.299 & 0.622 & 0.851 & 0.606 & 0.615 & 0.881 & 0.596 \\
\cmidrule(lr){2-15}
& \multirow{3}{*}{SVAMP}
& 10 & 0.057 & 0.309 & 0.133 & 0.170 & 0.332 & 0.160 & 0.169 & 0.797 & 0.169 & 0.171 & 0.765 & 0.169 \\
& & 20 & 0.061 & 0.483 & 0.133 & 0.176 & 0.486 & 0.160 & 0.170 & 0.797 & 0.169 & 0.171 & 0.798 & 0.169 \\
& & 30 & 0.066 & 0.490 & 0.133 & 0.181 & 0.486 & 0.160 & 0.170 & 0.932 & 0.169 & 0.171 & 0.929 & 0.169 \\

\midrule

\multirow{9}{*}{Qwen3-8B-Thinking}
& \multirow{3}{*}{GSM8K}
& 10 & 0.006 & 0.070 & 0.039 & 0.018 & 0.027 & 0.040 & 0.042 & 0.042 & 0.042 & 0.042 & 0.042 & 0.042 \\
& & 20 & 0.009 & 0.070 & 0.039 & 0.021 & 0.105 & 0.040 & 0.042 & 0.042 & 0.042 & 0.042 & 0.042 & 0.042 \\
& & 30 & 0.007 & 0.138 & 0.039 & 0.021 & 0.174 & 0.040 & 0.042 & 0.042 & 0.042 & 0.042 & 0.042 & 0.042 \\
\cmidrule(lr){2-15}
& \multirow{3}{*}{GSMHard}
& 10 & 0.144 & 0.781 & 0.169 & 0.128 & 0.793 & 0.163 & 0.190 & 0.190 & 0.190 & 0.190 & 0.190 & 0.190 \\
& & 20 & 0.144 & 0.842 & 0.169 & 0.128 & 0.833 & 0.163 & 0.190 & 0.190 & 0.190 & 0.191 & 0.191 & 0.190 \\
& & 30 & 0.144 & 0.854 & 0.169 & 0.128 & 0.824 & 0.163 & 0.190 & 0.191 & 0.190 & 0.190 & 0.192 & 0.190 \\
\cmidrule(lr){2-15}
& \multirow{3}{*}{SVAMP}
& 10 & 0.017 & 0.017 & 0.026 & 0.033 & 0.034 & 0.027 & 0.028 & 0.028 & 0.028 & 0.028 & 0.028 & 0.028 \\
& & 20 & 0.017 & 0.020 & 0.026 & 0.033 & 0.080 & 0.027 & 0.028 & 0.028 & 0.028 & 0.028 & 0.028 & 0.028 \\
& & 30 & 0.031 & 0.858 & 0.026 & 0.040 & 0.321 & 0.027 & 0.028 & 0.028 & 0.028 & 0.028 & 0.028 & 0.028 \\

\bottomrule
\end{tabular}%
}
\caption{
\textbf{Effect of bin count on calibration metrics.}
We report ECE, MCE, and Brier score for different bin counts across datasets and confidence estimators.
}
\label{tab:bin_ablation}
\end{table*}

\begin{figure*}[t]
\centering

\begin{minipage}[t]{0.49\textwidth}
\begin{tcolorbox}[
    enhanced,
    colback=gray!4,
    colframe=gray!55,
    boxrule=0.4pt,
    arc=2pt,
    left=2pt,
    right=2pt,
    top=2pt,
    bottom=2pt,
    boxsep=1pt,
    width=\linewidth,
    title=\textbf{Standard $\ptrue$}
]
\begin{lstlisting}[style=promptstyle]
<|begin_of_text|><|start_header_id|>system<|end_header_id|>

You are a helpful, respectful, and honest assistant. Always answer as helpfully as possible, while being safe. ...

<|start_header_id|>user<|end_header_id|>
Question:
Mary is baking a cake. The recipe calls for 6 cups of flour, 8 cups of sugar,
and 7 cups of salt. She already put in 5 cups of flour.
How many more cups of sugar than cups of salt does she need to add now?

Proposed answer:

To solve this problem, let's break it down step by step:
1. The recipe calls for 8 cups of sugar and 7 cups of salt.
2. Difference = 8 - 7 = 1.

#### 1

Is the proposed answer:
A: True
B: False
You must answer with exactly one character: either `A` or `B`.

The proposed answer is:


\end{lstlisting}
\end{tcolorbox}
\end{minipage}
\hfill
\begin{minipage}[t]{0.49\textwidth}
\begin{tcolorbox}[
    enhanced,
    colback=gray!4,
    colframe=gray!55,
    boxrule=0.4pt,
    arc=2pt,
    left=2pt,
    right=2pt,
    top=2pt,
    bottom=2pt,
    boxsep=1pt,
    width=\linewidth,
    title=\textbf{In-situ $\ptrue$}
]
\begin{lstlisting}[style=promptstyle]
<|begin_of_text|><|start_header_id|>system<|end_header_id|>

You are a helpful, respectful, and honest assistant. Always answer as helpfully as possible, while being safe. ...

<|start_header_id|>user<|end_header_id|>

You are a mathematics expert. First, study this example:
Question: Sarah has 10 marbles, gives 3 away.
Answer: Sarah would have 10 - 3 = <<10-3=7>>7 marbles remaining.
#### 7

Next, solve:
Mary is baking a cake. The recipe calls for 6 cups of flour, 8 cups of sugar,
and 7 cups of salt. She already put in 5 cups of flour.
How many more cups of sugar than cups of salt does she need to add now?

At the end, write the answer as an integer after '#### '.

<|start_header_id|>assistant<|end_header_id|>

To solve this problem, let's break it down step by step:
...
#### 1

Is the proposed answer:
A: True
B: False
You must answer with exactly one character: either `A` or `B`.

The proposed answer is:
\end{lstlisting}
\end{tcolorbox}
\end{minipage}

\caption{
\textbf{Standard $\ptrue$ vs. in-situ $\ptrue$} on GSM8K with Llama-3.1-8B.
Left: standard $\ptrue$ self-verification via re-prompting, where the full question--answer pair is re-encoded. Right: in-situ $\ptrue$, where the verification prompt is appended directly to the original generation trajectory and only the appended verification tokens require additional encoding.
}

\label{fig:prompt_comparison}
\end{figure*}

\section{Licensing of Artifacts}
\label{app:license}
Our experiments use GSM8K~\citep{cobbe2021training}, GSMHard~\citep{gao2022pal}, and SVAMP~\citep{patel2021nlp}, three publicly available mathematical question-answering benchmarks released for research use. Our BALD-score computation is adapted from the publicly available BAAL implementation~\citep{atighehchian2019baal}, which is distributed under the Apache License 2.0. To support reproducibility, we will release our implementation under the GPL-3.0 license.

\section{Use of AI Assistants}
\label{app:ai-assistants}

We used LLMs, including GPT, only to polish the writing (e.g., grammar and phrasing). All research ideas, methods, experiments, and substantive content were conceived and written by the authors.

% \section{Example Appendix}
% \label{sec:appendix}

% This is an appendix.

\end{document}